\documentclass{article}
\usepackage{amsmath}
\usepackage[preprint]{corl_2026} 
\usepackage{graphicx}
\usepackage{amssymb}
\usepackage{booktabs}
\graphicspath{{figures/}} 

\usepackage{xcolor}
\definecolor{deepteal}{RGB}{0, 102, 102}
\definecolor{MyBlue}{RGB}{24, 75, 153}

\title{MB-Loc: \textcolor{MyBlue}{M}ulti-planar \textcolor{MyBlue}{B}ird's-eye-view \textcolor{MyBlue}{Loc}alization in outdoor LiDAR scenes}

\author{
  Ayaan Choudhury \quad Preet Savalia \quad Anirudh Pydah \quad Avinash Sharma \\[0.2cm]
  Indian Institute of Technology Jodhpur, India \\[0.2cm]
  \texttt{\{b23me1013, b22ai036, anirudhp, avinashsharma\}@iitj.ac.in}
}

\begin{document}
\maketitle

\begin{abstract}
Global LiDAR localization is a fundamental task for autonomous navigation systems. Recent methods perform Scene Coordinate Regression (SCR) and achieve superior accuracy over Absolute Pose Regression (APR) solutions by predicting dense 3D world coordinates. However, SCR approaches introduce two major bottlenecks: severe computational inefficiency from processing raw 3D geometries and significant performance degradation under varying sensor viewpoints. To address these limitations, we present MB-Loc, a lightweight and viewpoint-robust SCR framework. Instead of relying on heavy 3D convolutions, we project the input LiDAR scan into a 2.5D Multi-planar Bird's-Eye View (BEV) representation. By slicing the point-cloud along the Z-axis and mapping signed depths into discrete 2D planes, MB-Loc retains essential 3D geometric structures while exploiting the computational tractability of standard 2D CNNs. To handle the inherent sparsity of outdoor LiDAR, we introduce a KL-regularized latent bottleneck that explicitly models spatial uncertainty without injecting stochastic noise. Finally, to ensure rotation robustness, we apply 3D spatial augmentations prior to planar projection, forcing the network to implicitly learn viewpoint-invariant features. We perform extensive experiments on the publicly available NCLT dataset and demonstrate that our proposed method outperforms the current state-of-the-art. Operating at real-time inference speeds, MB-Loc significantly outperforms traditional 3D-SCR architectures in computational efficiency.
\end{abstract}

\keywords{Autonomous driving and flight, Learning representations for robotic perception and control, Scene Coordinate Regression, Multi-plane BEV} 

\section{Introduction}
\label{sec:intro}
A robust localization task enables a robot to find its 6-Degree-of-Freedom (DoF) pose given a query (local) observation and a pre-built global reference map, thus serving as the critical anchor for any autonomous navigation system. 
In the context of LiDAR-based sensing, the objective is to estimate a precise 6-DoF robot pose in a 3D reference (global) point-cloud map using an input (local) point-cloud captured at a specific time.
There are multiple challenges associated with the aforementioned task, namely, under-sampling of smaller 3D structures in the scene, LiDAR sensor noise and beam divergence behavior, dynamic nature of scenes, severe view-specific occlusions in sensing and often redundant sampling in certain areas like ground points. Additionally, faster inference is critical in real-time navigation scenarios. 

Traditionally, global LiDAR localization methods~\cite{5152473, 9009829} relied on handcrafted or 
learned local feature descriptors to establish 3D correspondences between query scans and pre-built reference maps, followed by RANSAC-based~\cite{10.1145/358669.358692} pose estimation. However, these methods are highly susceptible to sensor noise, occlusions and often have slow inference time.   
Recently, learning-based methodologies have dominated this domain, where the two key directions are Absolute Pose Regression (APR)~\cite{li2024diffloc, wang2023hypliloc, 9928031, 10296854, wang2021pointlocdeepposeregressor, YU2022108685} and Scene Coordinate Regression (SCR)~\cite{li2023sgloc, 10654844, li2025lightloc}. The former attempts to directly map the high-dimensional point-clouds to a low-dimensional global pose vector through a deep neural network. While favored for their compact architectures and low-latency inference, the APR approach is fundamentally ill-posed as it attempts to memorize point-cloud coordinates. 

 \begin{figure}[t]
\centering
\includegraphics[width=0.7\linewidth, trim={0.2cm 0.35cm 0 0.2cm}, clip]{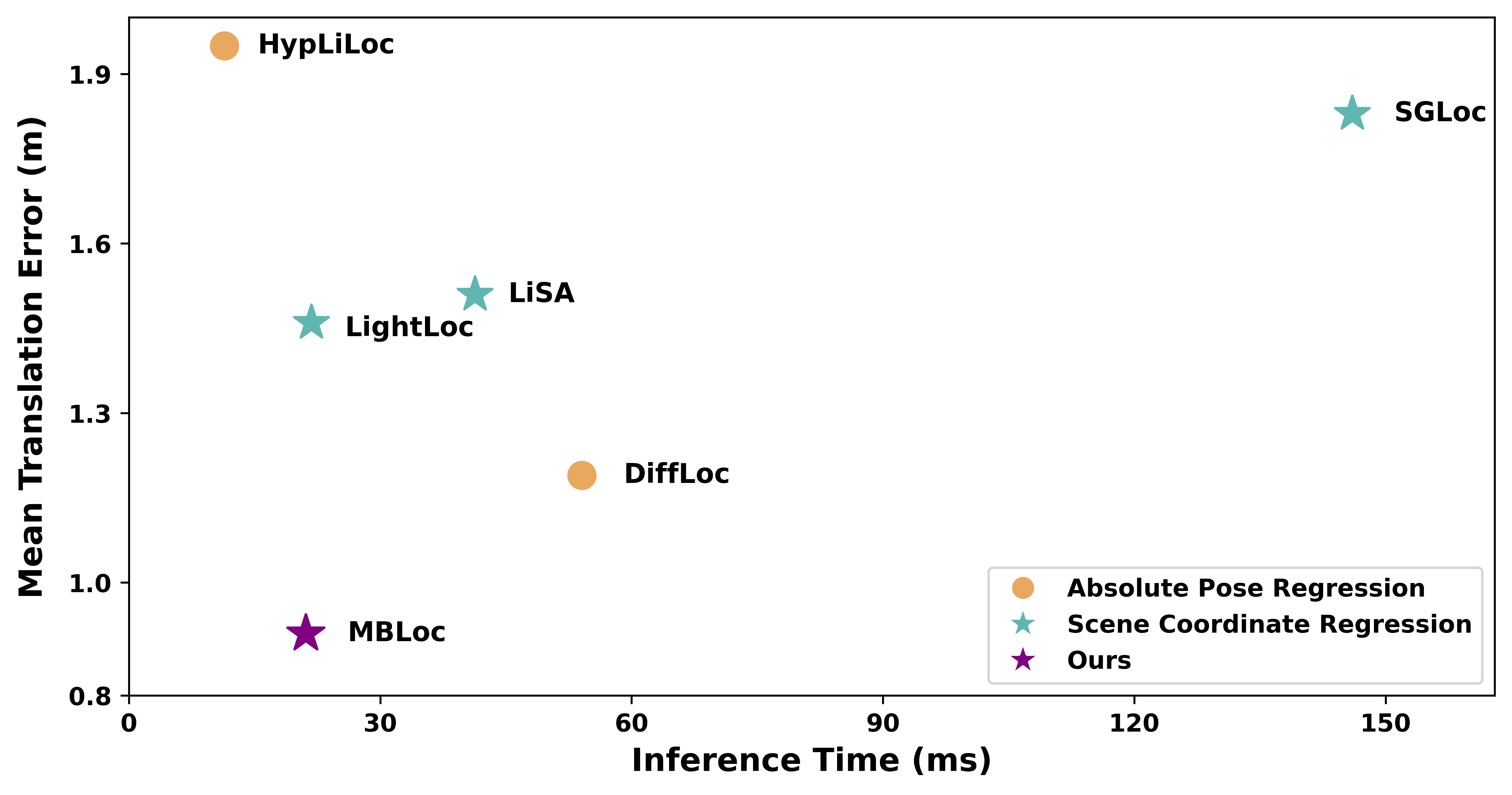}
   \caption{LiDAR localization performance vs. inference trade-off on NCLT~\cite{article} dataset.}
\label{fig:teaser}
\vspace{-0.5cm}
\end{figure}

Conversely, the latter (i.e., SCR) directly regresses the 3D world coordinates for input query scans and subsequently establishes explicit 3D-to-3D geometric correspondences between the current observation scan and the global reference map. 
However, contemporary SCR methods introduce a major operational challenge: computational inefficiency. To preserve the complex geometric structures of outdoor environments, SCR networks rely heavily on 3D sparse convolutions or heavy point-based architectures. A common alternative to reduce dimensionality is to project the 3D point-clouds into 2D planes. However, standard 2D Bird's-Eye-View (BEV) projections suffer from severe top-down occlusion, where higher structures like tree canopies overwrite crucial ground-level features like curbs, resulting in catastrophic geometric information loss and degraded localization accuracy.

In this work, we propose \textbf{MB-Loc}, a novel and lightweight SCR framework that bridges the gap between computational efficiency and geometric fidelity. First, we propose a novel efficient representation for LiDAR point-cloud scan. Instead of relying on expensive 3D architectures or using a single 2D BEV representation (with significant quantization), we encode the input LiDAR scan into a 2.5D Multi-planar BEV representation. This is achieved by slicing the point-cloud along the Z-axis into multiple (discrete horizontal) planes (named as  \textit{Z-slicing}) and then mapping 3D points to these planes by storing the respective signed depths into associated 2D grids, yielding a sparse set of projected LiDAR points. This multi-planar BEV representation helps to address the challenge of top-down occlusion to retain the essential 3D geometric structures of the environment while attaining the computational efficiency using the standard 2D CNN architecture to encode them. Thereafter, by integrating dual spatial and channel attention mechanisms, our method explicitly learns to isolate and leverage the most distinctive geometric features across the Z-slicing planes. 

Subsequently, we introduce a latent bottleneck inspired by Information Bottleneck principles~\cite{alemi2019deepvariationalinformationbottleneck, tishby2015deep} where instead of propagating raw, fragmented feature maps, the network learns a representation of structural uncertainty parameterized by $\mu$ and $\sigma$. While leveraging a KL-divergence~\cite{kullback1951information} penalty commonly associated with stochastic VAEs~\cite{kingma2013auto}, we bypass stochastic sampling entirely, opting instead for a deterministic projection modulated by an adaptively learned scale factor. By regularizing this latent space and dynamically scaling the encoded uncertainty, we prevent the architecture from overfitting to discrete, noisy point returns. Next, the network predicts 3D coordinate offsets that are added to the projected LiDAR points in Z-sliced Multi-planar BEV 2D grids to reconstruct the dense 3D world coordinates. Finally, these explicit correspondences are then passed to a closed-form Singular Value Decomposition (SVD) RANSAC backend, which effectively rejects residual quantization noise to recover a highly precise global pose.
Figure~\ref{fig:teaser} shows that our MB-Loc achieves the lowest error with a competitive inference latency.

\section{Related Work}
\label{sec:related}
Global LiDAR localization fundamentally bifurcates into Absolute Pose Regression (APR) and Scene Coordinate Regression (SCR). APR 
methods~\cite{wang2021pointlocdeepposeregressor, wang2023hypliloc, 
li2024diffloc} directly estimate 6-DoF poses; while theoretically compact, 
they often lack the explicit geometric priors necessary for centimeter-level 
precision, and iterative variants~\cite{10296854} introduce severe inference latency. Conversely, SCR methods reframe localization as a dense correspondence task: SGLoc~\cite{li2023sgloc} decouples correspondence regression from pose calculation via RANSAC~\cite{10.1145/358669.358692}, 
LiSA~\cite{10654844} enriches geometric manifolds through semantic 
distillation, and LightLoc~\cite{li2025lightloc} targets improved efficiency, yet all remain bottlenecked by heavy 3D sparse convolutions or complex point-based structures.

Spatial representation heavily dictates this accuracy-latency trade-off. 3D methods preserve geometric detail at high computational cost, whereas standard 2D BEV or range image projections~\cite{Chen_2020} unlock 2D CNN efficiency but suffer from severe Z-axis occlusion. Attention 
mechanisms~\cite{Ma_2022} partially mitigate this projection loss by 
selectively emphasizing informative structural regions, but cannot recover 
discarded vertical geometry. Descriptor-based methods like 
DiSCO~\cite{9359460} and Scan Context++~\cite{kim2021scan} utilize polar or layered Cartesian representations, but are engineered exclusively for place recognition; their spatial pooling mechanisms irreparably destroy the 
exact metric point identities required for dense 3D-to-3D  correspondences in 
SCR. Alternatively, layered 2D representations have demonstrated effective geometric preservation without 3D convolutions in adjacent domains: \cite{single_view_mpi} represent scenes as depth-discretized 
fronto-parallel RGBA plane stacks for novel view synthesis, while \cite{jinka2022sharpshapeawarereconstructionpeople} encode clothed 
human geometry as successive ray-traced depth peel maps for monocular 3D reconstruction, both motivating our Z-sliced Multi-planar design.

\section{Proposed MB-Loc Method}
\label{sec:method}
\begin{figure}[ht]
\begin{center}
    \makebox[\textwidth][c]{%
  \fbox{\includegraphics[width=1\linewidth]{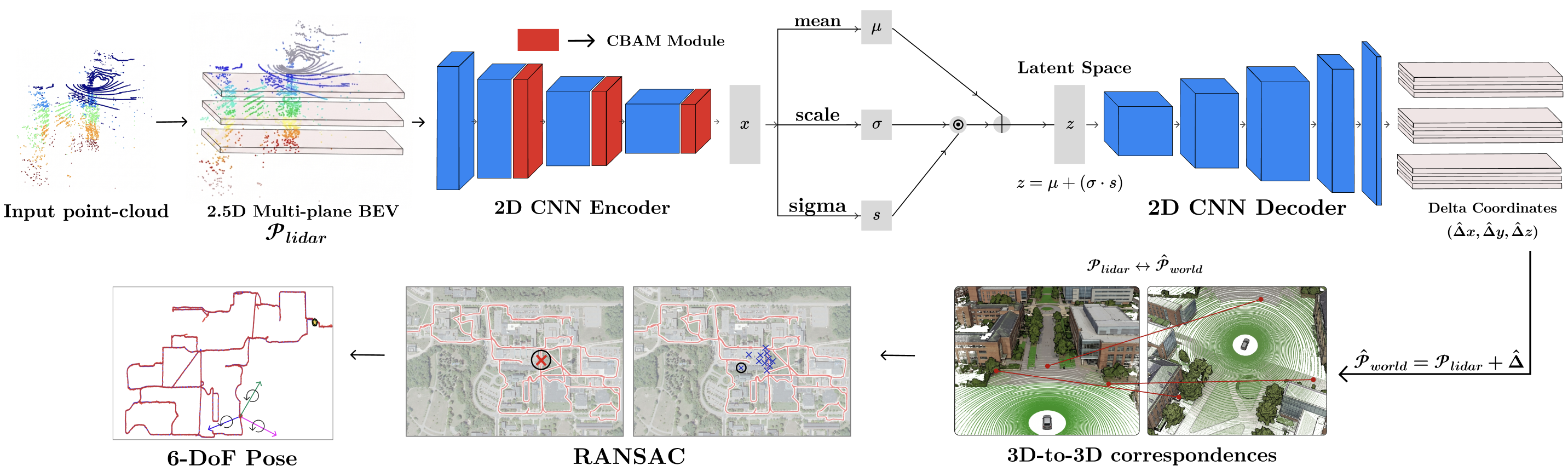}}}
\end{center}
\caption{Overview of the proposed MB-Loc during inference.} 

\label{fig:architecture}
\end{figure}

Figure~\ref{fig:architecture} depicts an overview of the proposed MB-Loc method, where the input 3D point-cloud is sliced along the Z-axis to recover a 2.5D Multi-planar BEV representation.
Next, a 2D CNN (ResNet) Encoder with CBAM~\cite{woo2018cbamconvolutionalblockattention} extracts geometric features across the Z-slicing planes. 
The feature map $x$ then enters a regularized deterministic bottleneck where parallel projection heads yield mean ($\mu$), standard deviation ($\sigma$), and scale factor ($s$), combined via $z = \mu + \sigma \cdot s$.
Thereafter, a 2D CNN Decoder inflates $z$ to native resolution, regressing dense coordinate deltas $(\hat{\Delta}x, \hat{\Delta}y, \hat{\Delta}z)$ per plane. 
Finally, adding these offsets to the input coordinates reconstructs the dense 3D world points, establishing dense 3D-to-3D  correspondences for the RANSAC solver to recover the global 6-DoF pose.

\subsection{2.5D Multi-planar BEV Representation}
\label{subsec:projection} 
Let the input point-cloud be $\mathcal{P} = \{p_i \in \mathbb{R}^3 \mid i = 1 \dots N\}$, where $p_i = (x_i, y_i, z_i)$. We bound the spatial extent of the point-cloud based on the dynamic minimum and maximum extents along each axis: $[x_{min}, x_{max}]$, $[y_{min}, y_{max}]$, and $[z_{min}, z_{max}]$.

We partition the Z-axis into $P$ discrete horizontal Z-slicing planes of uniform thickness $w_z = \frac{z_{max} - z_{min}}{P}$. Each point $p_i$ is assigned a plane index $k \in \{0, \dots, P-1\}$, computed sequentially from the maximum Z-elevation downward:
\begin{equation}
    k = \text{clamp}\left(\left\lfloor \frac{z_{max} - z_i}{w_z} \right\rfloor, 0, P-1\right)
\end{equation}
Simultaneously, we map the $(x_i, y_i)$ spatial coordinates to a dense 2D grid $(u_i,v_i)$ of spatial resolution $G \times G$ where $u$ is defined as:
\begin{equation}
    u = \text{clamp}\left(\left\lfloor \frac{x_i - x_{min}}{x_{max} - x_{min}} \cdot (G - 1) \right\rfloor, 0, G-1\right).
\end{equation} 
Similarly, the $v_i$ coordinate is defined along the Y-axis. Within each Z-slicing plane $k$, we define a local depth reference $z_{start}^{(k)}$ corresponding to the lower bound of the slice: $z_{start}^{(k)} = z_{max} - (k + 1) \cdot w_z$. The signed relative depth for each point is then calculated as: $\delta z_i = z_i - z_{start}^{(k)} + \epsilon$, where $\epsilon$ is a small constant added for numerical stability. Because multiple points from the sparse point-cloud may project into the same $(k, u, v)$ grid point, we handle spatial collisions through a scatter-reduce operation, retaining the point that minimizes the absolute relative depth ($|\delta z_i|$).

This projection process yields the primary input tensor $\mathcal{V}$, which contains the projected LiDAR points and encodes the scene's vertical geometry. Next, we generate three additional dense representations based on the retained points. First, a binary validity mask $\mathcal{M}$ is created to distinguish occupied pixels from empty space: $\mathcal{M}(k, u, v) = 1$ if occupied, and $0$ otherwise. Second, the original spatial coordinates of the retained points are cached into a local coordinate tensor, $\mathcal{C}_{LiDAR}$. Third, the corresponding ground-truth global map coordinates for these specific points are stored in $\mathcal{C}_{world}$. The ultimate regression target for our network is the tensor of spatial offset, $\Delta$, formulated as the difference between the global world frame and the local LiDAR frame: $\mathcal{C}_{world} = \mathcal{C}_{LiDAR} + \Delta$.
By framing the objective around the prediction of bounded $\Delta$ offsets rather than absolute world coordinates, we significantly constrain the regression space, allowing the network to learn local structural alignments efficiently.

\subsection{Network Architecture}
\label{subsec:architecture}
Our network translates the 2.5D Multi-planar BEV representation into dense 3D coordinate offsets via a fully-convolutional encoder-decoder. The encoder utilizes residual blocks and Convolutional Block Attention Modules (CBAM)~\cite{woo2018cbamconvolutionalblockattention} to isolate informative Z-slicing planes, prioritizing structural details over empty space. However, outdoor LiDAR data inherently suffers from physical constraints such as range-dependent sparsity and beam divergence; these issues are further compounded by the geometric quantization artifacts introduced by our Z-slicing. To overcome these challenges, we introduce a deterministic latent bottleneck. Finally, the decoder upsamples this regularized latent state to predict dense point-wise 3D spatial offsets ($\hat{\Delta}$) (Eq.~\ref{eq:decoder_output}), providing the exact 3D-to-3D correspondences required by the RANSAC backend for global pose estimation.  Please refer to the appendix~\ref{sec:appendix_implementation} for more details.

\subsection{Training Objectives}
\label{subsec:loss}

During training, the network parameters are optimized by minimizing a composite loss function comprising a masked coordinate regression loss and a latent regularization term. 
To penalize geometric misalignment, we employ an L1 coordinate loss due to inherent sparsity in the representation. 
Additionally, the loss is strictly gated by the binary validity mask $\mathcal{M}$ to exclude the empty regions. The loss computes the mean absolute error exclusively between the predicted spatial offsets $\hat{\Delta}$ and the ground-truth offsets $\Delta$ at the occupied grid locations:
\begin{equation}
    L_{coord} = \frac{1}{\sum_{k,u,v} \mathcal{M}(k,u,v)} \sum_{k,u,v} \mathcal{M}(k,u,v) \left\| \Delta(k,u,v) - \hat{\Delta}(k,u,v) \right\|_1
\end{equation}

Simultaneously, to enforce structural generalization within the latent bottleneck, we apply a KL~\cite{kullback1951information} divergence penalty. This term coerces the predicted spatial distribution parameters ($\boldsymbol{\mu}$ and $\boldsymbol{\sigma}$) towards a standard normal prior $\mathcal{N}(0, \mathbf{I})$. For the latent feature maps possessing $C$ channels and spatial dimensions $H'$ and $W'$, the regularization is formulated as:
\begin{equation}
    L_{KL} = \frac{1}{2 C H' W'} \sum_{c=1}^{C} \sum_{h=1}^{H'} \sum_{w=1}^{W'} \left( \boldsymbol{\mu}_{c,h,w}^2 + \boldsymbol{\sigma}_{c,h,w}^2 - 1 - \log(\boldsymbol{\sigma}_{c,h,w}^2 + \gamma) \right)
\end{equation}
where $\gamma$ is a small constant preventing logarithmic instability.  Rather than treating the bottleneck as a stochastic channel, the KL term 
acts purely as a feature-smoothing regularizer, explicitly preventing 
$\boldsymbol{\sigma}$ from collapsing to zero and preserving the 
uncertainty-aware structure of the latent space. By penalising sharp, high-variance deviations from a unit normal prior, we compress the feature representation without requiring any sampling operation, preventing the network from memorizing transient noise or quantization artifacts.
Thus, the final training objective is:
$   L_{total} = L_{coord} + \lambda L_{KL}
$,
where $\lambda$ is a hyperparameter scaling the influence of the latent regularization relative to the geometric coordinate regression.
\subsection{Pose Optimization using RANSAC}
\label{subsec:pose}
Once the network infers $\hat{\Delta}$, we extract occupied source points 
$\mathcal{C}_{LiDAR}$ via the validity mask $\mathcal{M}$ and construct 
predicted global target points as $\hat{\mathcal{C}}_{world} = 
\mathcal{C}_{LiDAR} + \hat{\Delta}$, maintaining strict one-to-one 
correspondences for all retained structural points.
We employ a RANSAC-based~\cite{10.1145/358669.358692} estimator to recover 
the rigid transformation while filtering regression outliers and quantization 
noise. Each iteration samples $N=3$ correspondences and solves for the optimal 
rotation $\mathbf{R}$ (Eq.~\ref{eq:ransac_R}) and translation $\mathbf{t}$ (Eq.~\ref{eq:ransac_t}) via closed-form SVD using Kabsch's algorithm~\cite{1976AcCrA..32..922K}, computed from the sampled centroids and cross-covariance matrix (Eqs.~\ref{eq:ransac_centroids},~\ref{eq:ransac_H}). To maximize efficiency, 
iterations terminate early once the minimum count $k$, required to guarantee confidence $p$, is reached:
$
    k = log(1-p)/log(1-w^3)
$, where $w$ is the dynamically estimated inlier ratio under a predefined distance threshold~~$\tau$. The transformation with the maximum inlier consensus is selected, followed by a final least-squares SVD refinement over all inliers to recover the 6-DoF pose. We use 
$p=0.95$ in our experiments.
\section{Experiments \& Results}
\label{sec:experiments}

\subsection{Experimental Settings}
\label{subsec:exp_settings}

\textbf{Benchmark Dataset:} We evaluate the proposed MB-Loc framework on the NCLT dataset~\cite{article}. Following the standard evaluation protocol established by existing methods, we utilize the data from 2012-01-22, 2012-02-02, 2012-02-18, and 2012-05-11 as our training set. For evaluation, we test on the trajectories from 2012-02-12, 2012-02-19, 2012-03-31, and 2012-05-26. While the Oxford Radar RobotCar~\cite{barnes2020oxford} dataset is another common benchmark in this domain, we were unable to secure the necessary data access permissions prior to publication. Consequently, our extensive evaluations focus primarily on the highly challenging, long-term NCLT~\cite{article} benchmark. Nevertheless, to broaden our comparison, we additionally conduct experiments on a heavily subsampled variant of the Oxford~\cite{barnes2020oxford} dataset, which we report in the supplementary material (Section~\ref{sec:supp_oxford}).

\textbf{Implementation Details:} 
All experiments are implemented in PyTorch~\cite{paszke2019pytorch} and executed on a single NVIDIA A30 GPU. The network is trained using the Adam optimizer with an initial learning rate of $3 \times 10^{-3}$ and a weight decay of $1 \times 10^{-6}$. We employ a step learning rate scheduler, decaying the learning rate by a factor of $0.85$ every $20$ epochs. Furthermore, the KL regularization weight for the deterministic latent bottleneck is set to $1 \times 10^{-4}$. We apply random spatial augmentations during training. We rotate 80\% of the input LiDAR scans by a random yaw angle between -180$^{\circ}$ and 180$^{\circ}$. Additionally, 50\% of the scans are randomly shifted in the $xy$-plane by up to $\pm$2m in both the $x$ and $y$ directions. By applying these augmentations before the projection, we force the network to implicitly learn viewpoint-invariant structural representations. The core augmentation pipeline is adapted from the official open-source implementation of DiffLoc~\cite{li2024diffloc}.
We use the following configuration: $P=15$ Z-slicing planes, spatial resolution $G=512$, RANSAC threshold $\tau = 4.0$ m, and maximum correspondence subset size $N_{\max} = 2000$. 

\subsection{Localization Results}
\begin{figure*}[t]
\begin{center}
    \makebox[\textwidth][c]{%
        \begin{minipage}[b]{0.16\textwidth}
            \centering
            \includegraphics[width=\linewidth]{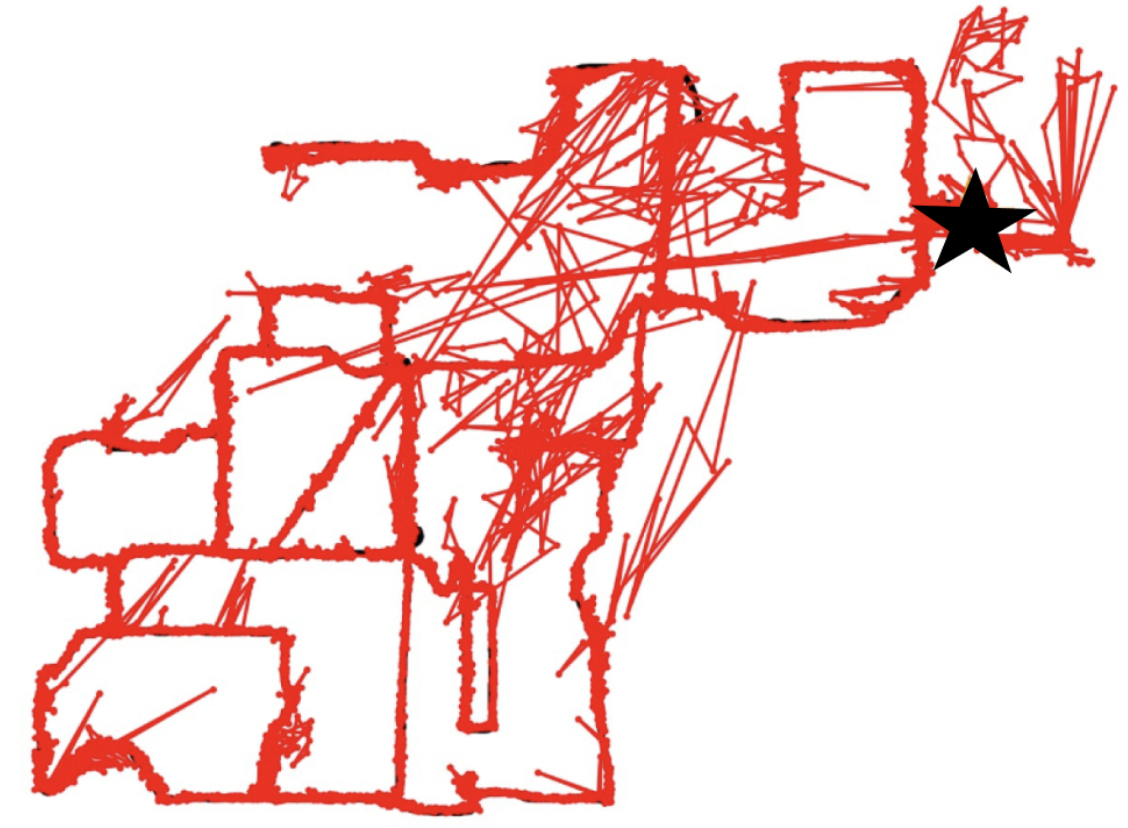}
            \small (a) NIDALoc (4.47m/3.53$^{\circ}$)
        \end{minipage}\hfill
        \begin{minipage}[b]{0.16\textwidth}
            \centering
            \includegraphics[width=\linewidth]{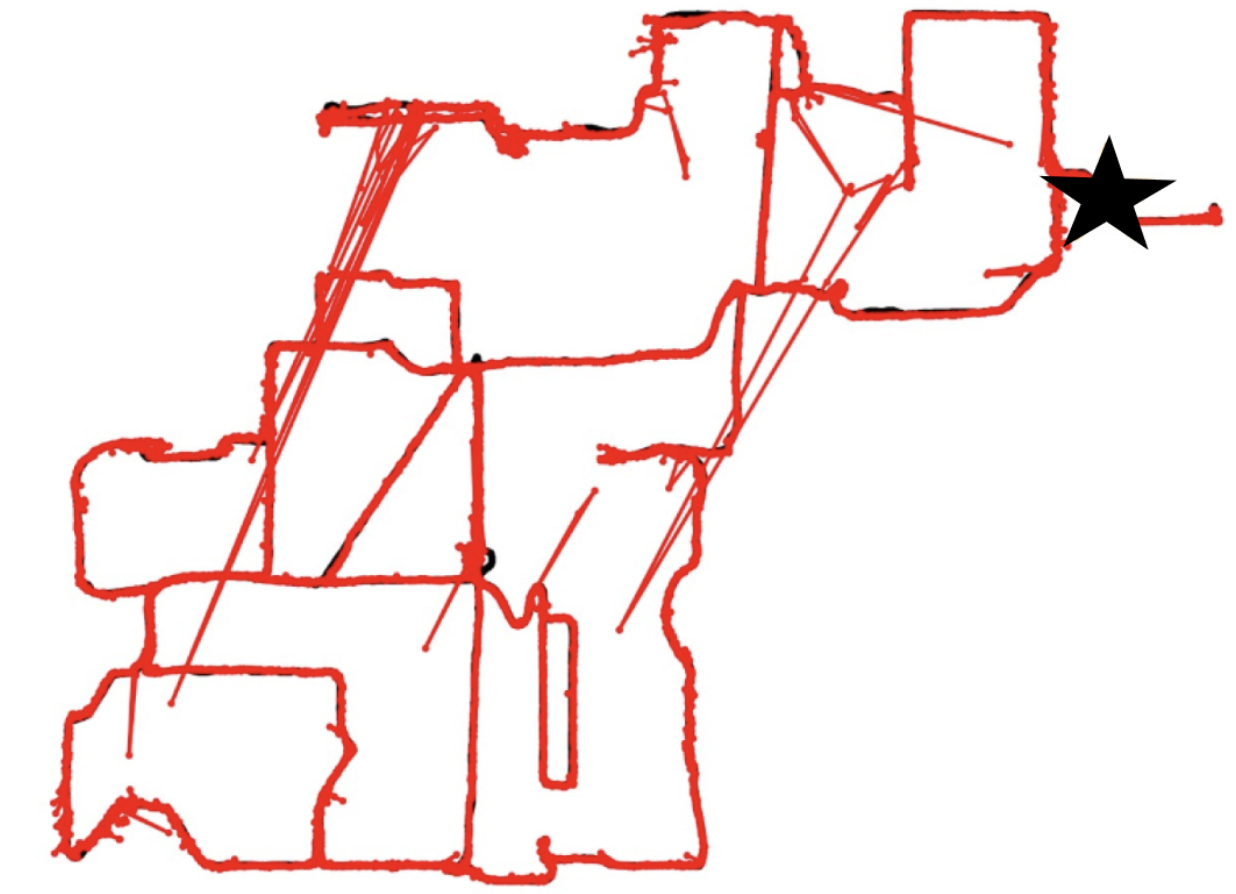}
            \small (b) HypLiLoc (1.95m/3.16$^{\circ}$)
        \end{minipage}\hfill
        \begin{minipage}[b]{0.16\textwidth}
            \centering
            \includegraphics[width=\linewidth]{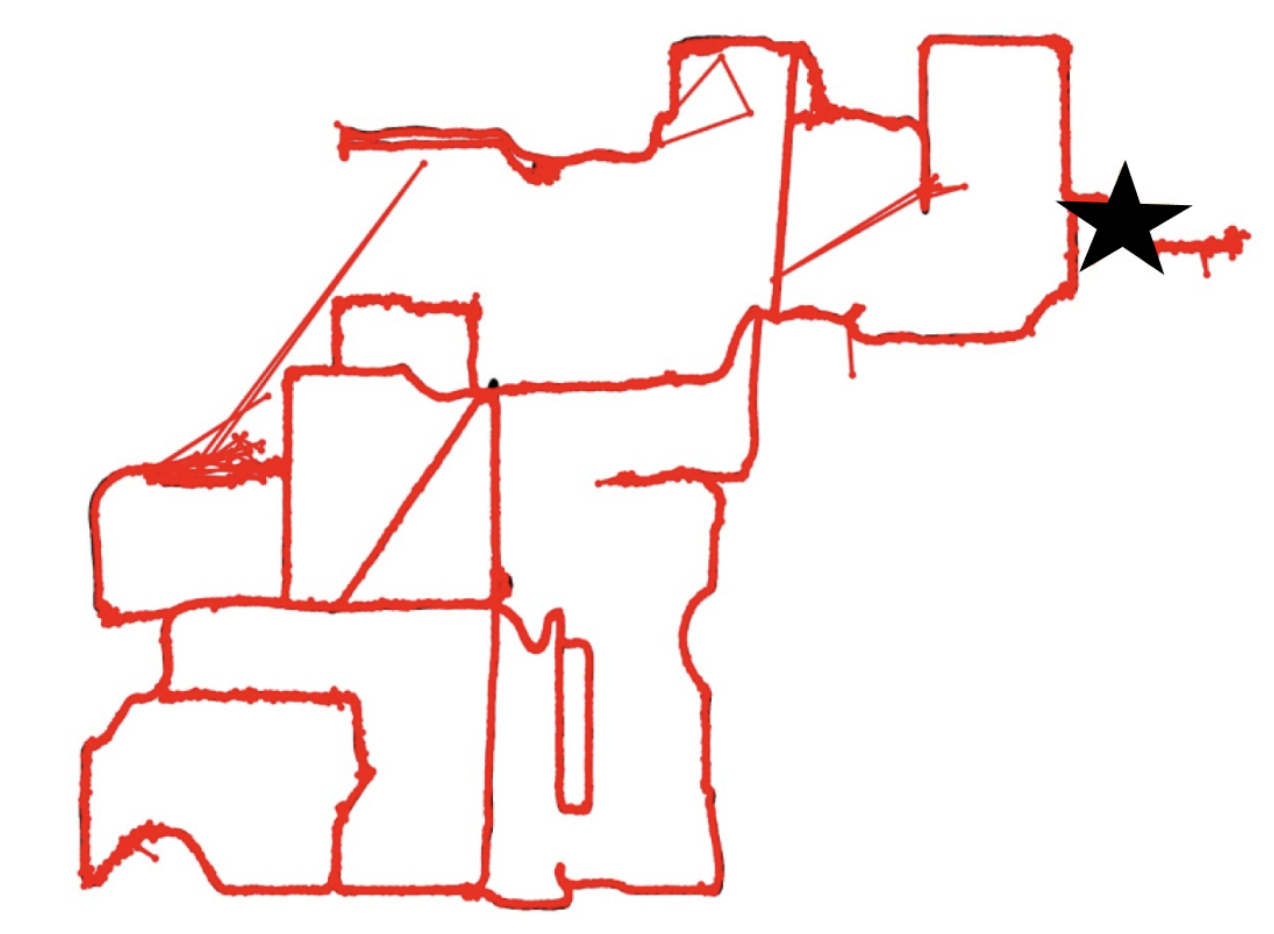}
            \small (c) SGLoc (1.83m/3.54$^{\circ}$)
        \end{minipage}\hfill
        \begin{minipage}[b]{0.16\textwidth}
            \centering
            \includegraphics[width=\linewidth]{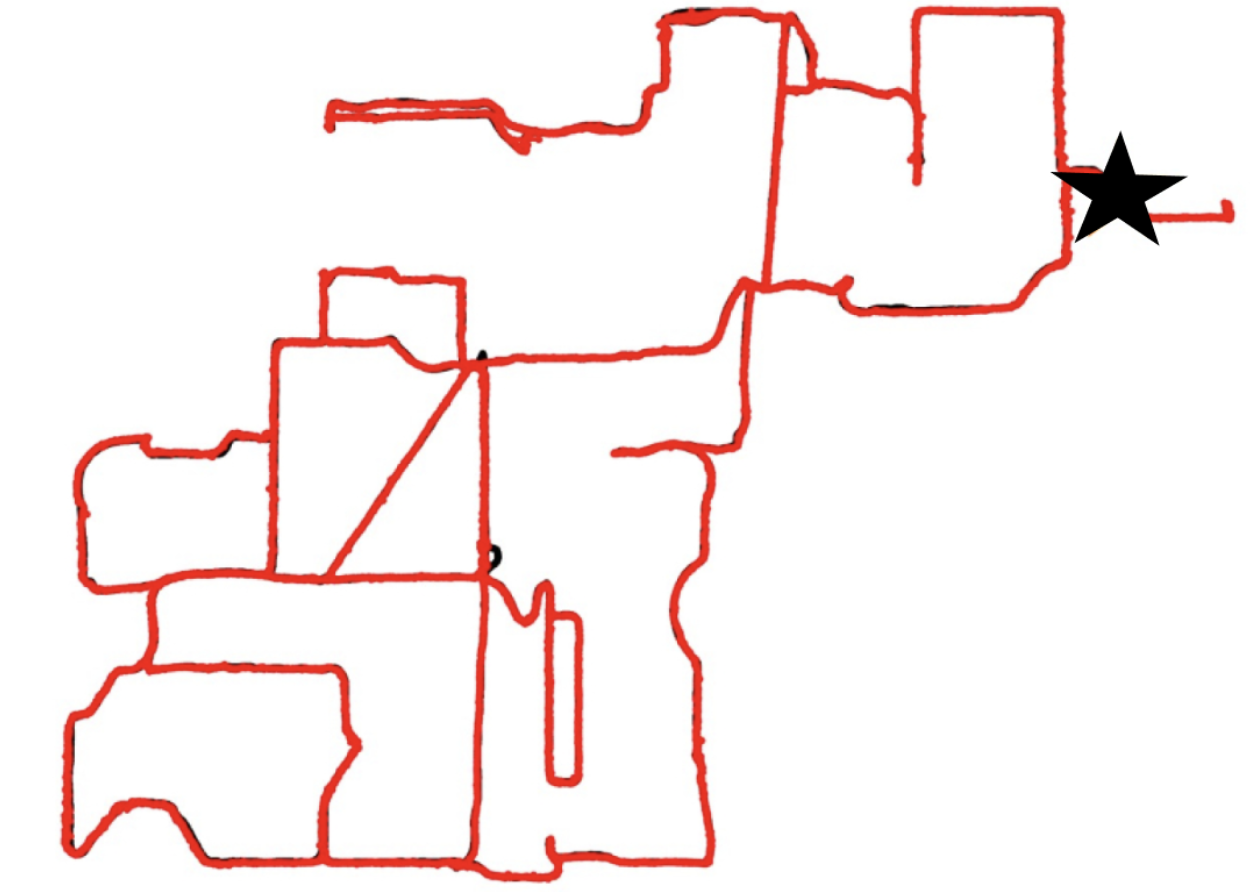}
            \small (d) DiffLoc (1.19m/2.31$^{\circ}$)
        \end{minipage}\hfill
        \begin{minipage}[b]{0.16\textwidth}
            \centering
            \includegraphics[width=\linewidth]{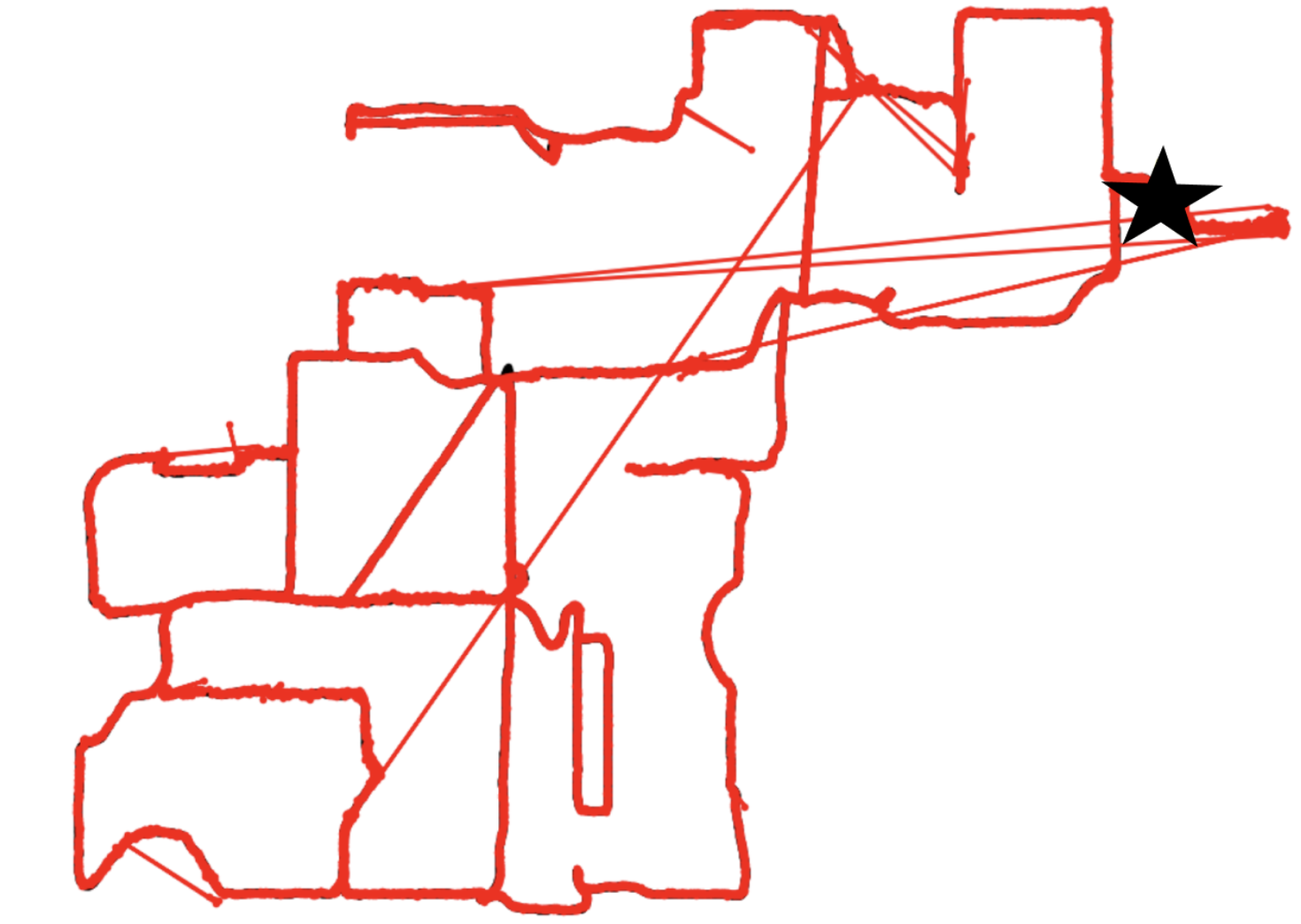}
            \small (e) LightLoc (1.46m/2.80$^{\circ}$)
        \end{minipage}\hfill
        \begin{minipage}[b]{0.16\textwidth}
            \centering
            \includegraphics[width=\linewidth]{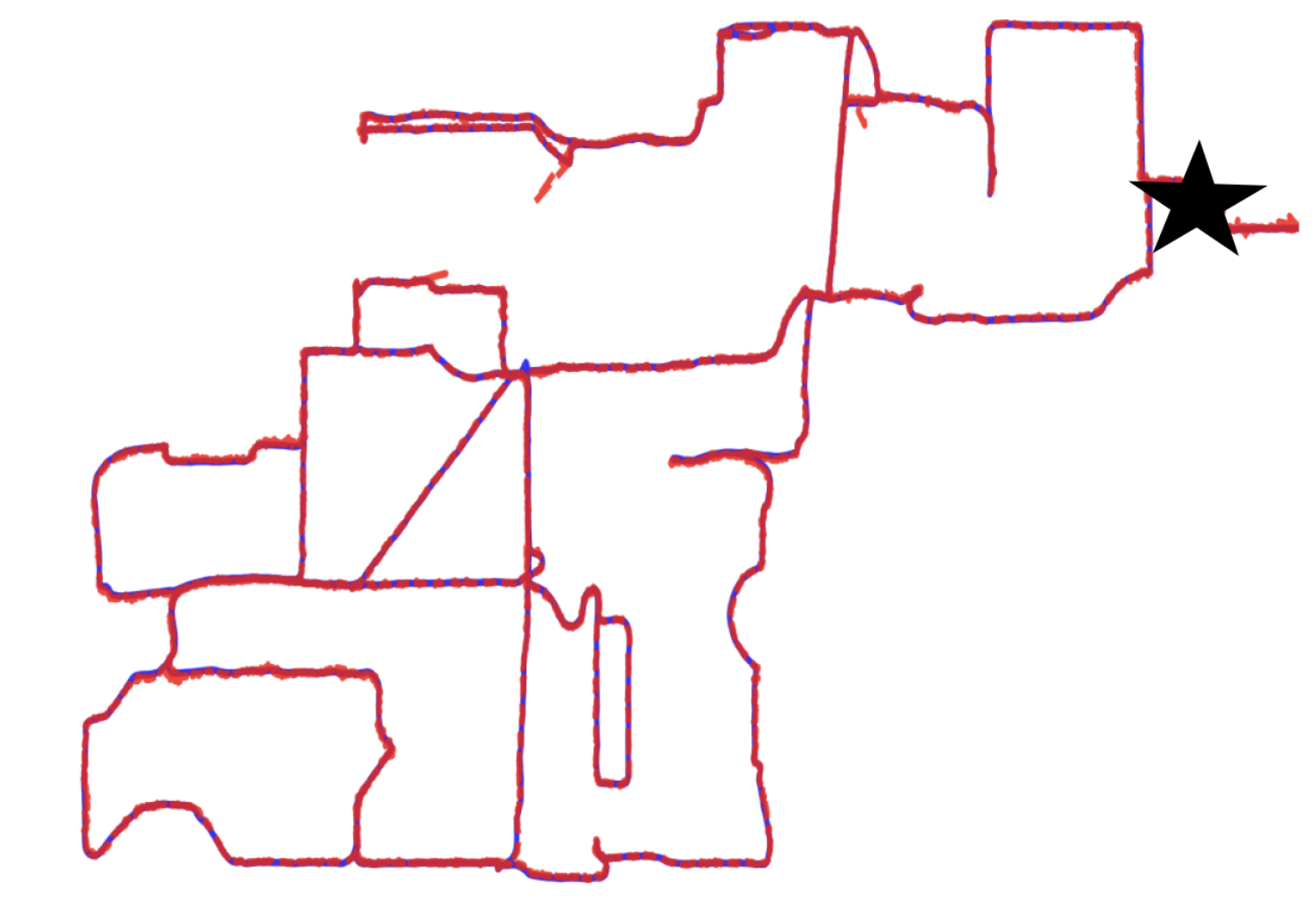}
            \small (f) MB-Loc (0.91m/2.30$^{\circ}$)
        \end{minipage}%
    }
\end{center}
\caption{Plot of predicted (red) vs.\ ground 
truth (black) trajectories.
$\bigstar$ denotes the first frame.}
\label{fig:trajectory_comparison_nclt}
\end{figure*}

\begin{figure}[h]
\begin{center}
   \begin{minipage}{0.48\linewidth}
      \centering
      \includegraphics[width=\linewidth]{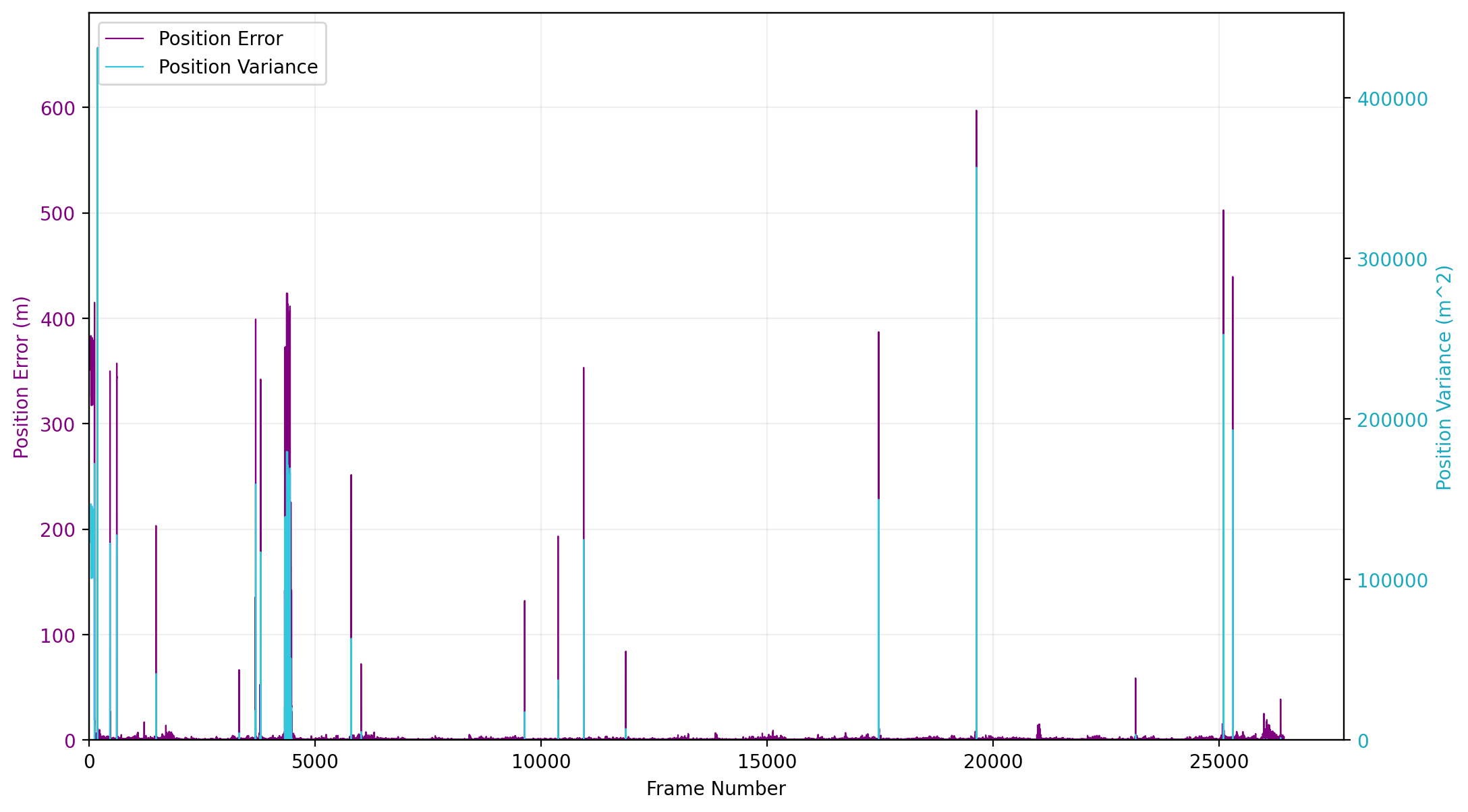}
      \centerline{{\small (a) LightLoc}}
   \end{minipage}
   \hfill
   \begin{minipage}{0.48\linewidth}
      \centering
      \includegraphics[width=\linewidth]{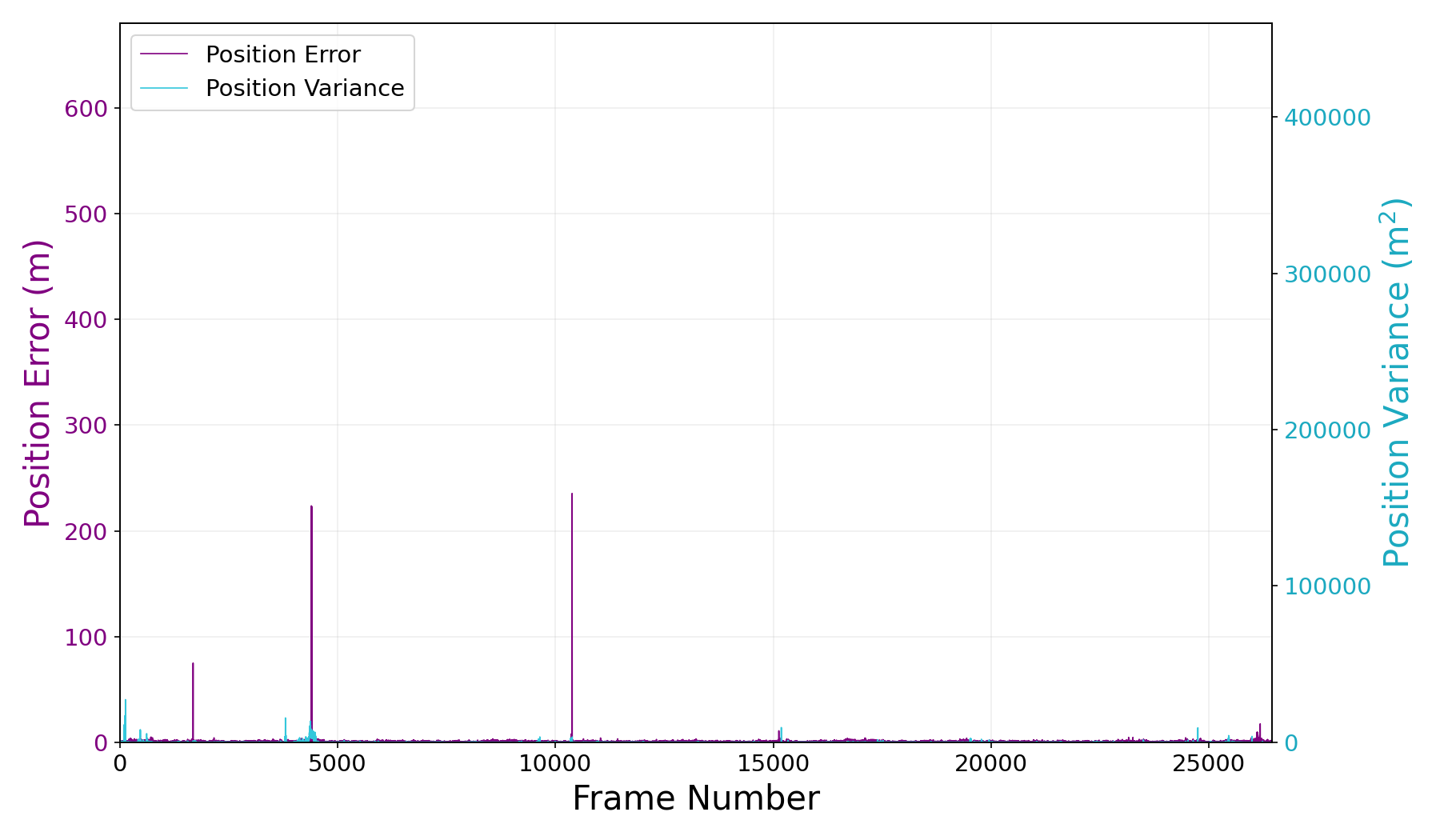}
      \centerline{{\small (b) MB-Loc (Ours)}}
   \end{minipage}
   \caption{Pose uncertainty evaluation results where we show the relationship between position error and variance on the 2012-05-26 trajectory of the NCLT~\cite{article} dataset.}
   \label{fig:uncertainty}
\end{center}
\end{figure}

The quantitative localization accuracy, measured by mean translation error (in meters) and mean rotation error (in degrees), is presented in Table~\ref{tab:nclt_results}. ``S'' denotes SCR methods while ``A'' denotes APR methods. MB-Loc achieves state-of-the-art performance with an average error of 0.91m and 2.30$^{\circ}$, demonstrating superior translation accuracy and highly competitive rotational precision compared to existing baselines. Notably, MB-Loc consistently achieves sub-meter translation accuracy across all evaluation sequences, demonstrating robust generalization across seasonal variations in contrast to competing baselines, which 
exhibit substantially degraded performance on the 2012-05-26 sequence 
relative to their results on other sequences. Figure~\ref{fig:uncertainty} shows the position error and position 
variance of LightLoc and MB-Loc on the 2012-05-26 sequence. LightLoc 
exhibits frequent high-magnitude error spikes throughout the trajectory, 
with several outliers exceeding 500~m, alongside large variance spikes 
that are poorly correlated with the actual errors. In contrast, MB-Loc 
produces substantially fewer outliers with significantly lower peak 
magnitudes, demonstrating more stable and reliable localization across the full trajectory. 
In Figure~\ref{fig:trajectory_comparison_nclt}, we show a comparison of predicted trajectories against ground truth on the 2012-02-12 sequence. MB-Loc produces the closest alignment to 
the ground truth path, with no visible large-scale deviations, in 
contrast to competing methods which exhibit varying degrees of 
trajectory drift and localization outliers.

\begin{table}[h]
\begin{center}
\caption{Quantitative results on the NCLT~\cite{article} dataset. We highlight the \textcolor{blue}{best} and \textcolor{magenta}{second best} results.}
\label{tab:nclt_results}
\small
\begin{tabular}{|l|c|c|c|c|c|c|}
\hline
\textbf{Method} & \textbf{Type} & \textbf{2012-02-12} & \textbf{2012-02-19} & \textbf{2012-03-31} & \textbf{2012-05-26} & \textbf{Avg [m/$^{\circ}$]} \\
\hline\hline
PNVLAD~\cite{uy2018pointnetvlad} & S & 7.75/6.49 & 7.47/5.49 & 6.98/5.67 & 14.34/7.93 & 9.14/6.40 \\
DCP~\cite{wang2019deep} & S & 9.84/6.84 & 8.27/5.16 & 8.94/5.96 & 15.62/7.99 & 10.67/6.49 \\
SGLoc~\cite{li2023sgloc} & S & 1.20/3.08 & 1.20/3.05 & 1.12/3.28 & 3.81/4.74 & 1.83/3.54 \\
LightLoc~\cite{li2025lightloc} & S  & 0.98/2.76 & \textcolor{magenta}{0.89}/2.51 & \textcolor{blue}{0.86}/2.67 & 3.10/3.26 & 1.46/2.80 \\
LiSA~\cite{10654844} & S & \textcolor{magenta}{0.97}/\textcolor{blue}{2.23} & 0.91/\textcolor{blue}{2.09} & \textcolor{magenta}{0.87}/\textcolor{blue}{2.21} & 3.30/2.84 & 1.51/2.34 \\
PointLoc~\cite{wang2021pointlocdeepposeregressor} & A & 7.23/4.88 & 6.31/3.89 & 6.71/4.32 & 10.02/5.32 & 7.57/4.60 \\
PosePN~\cite{YU2022108685} & A & 9.45/7.47 & 6.15/5.05 & 5.79/5.28 & 13.47/7.77 & 8.72/6.39 \\
PosePN++~\cite{YU2022108685} & A & 4.97/3.75 & 3.68/2.65 & 4.35/3.38 & 9.59/4.49 & 5.65/3.57 \\
PoseMinkLoc~\cite{YU2022108685} & A & 6.24/5.03 & 4.87/3.94 & 4.23/4.03 & 10.32/6.52 & 6.42/4.88 \\
PoseSOE~\cite{YU2022108685} & A & 13.09/8.05 & 6.16/4.51 & 5.24/4.56 & 12.60/7.67 & 9.27/6.20 \\
STCLoc~\cite{9928031} & A & 4.91/4.34 & 3.25/3.10 & 3.75/4.04 & 8.67/5.23 & 5.15/4.18 \\
NIDALoc~\cite{10296854} & A & 4.48/3.59 & 3.14/2.52 & 3.67/3.46 & 6.60/4.56 & 4.47/3.53 \\
HypLiLoc~\cite{wang2023hypliloc} & A & 1.71/3.56 & 1.68/2.69 & 1.52/2.90 & 2.90/3.47 & 1.95/3.16 \\
DiffLoc~\cite{li2024diffloc} & A & 0.99/2.40 & 0.92/\textcolor{magenta}{2.14} & 0.98/\textcolor{magenta}{2.27} & \textcolor{magenta}{1.88}/\textcolor{magenta}{2.43} & \textcolor{magenta}{1.19}/\textcolor{magenta}{2.31} \\
\hline
MB-Loc (Ours) & S & \textcolor{blue}{0.96}/\textcolor{magenta}{2.37} & \textcolor{blue}{0.88}/2.22 & 0.89/2.40 & \textcolor{blue}{0.93}/\textcolor{blue}{2.23} & \textcolor{blue}{0.91}/\textcolor{blue}{2.30} \\
\hline
\end{tabular}
\end{center}
\end{table}

To rigorously evaluate computational efficiency, we compare the runtime latency and memory footprint of MB-Loc against leading baselines in Table~\ref{tab:sota_efficiency} (where the batch size was set to 1 for each case).
\begin{table}[h]
\begin{center}
\caption{Algorithmic footprint and single-frame processing latency breakdown. Profiling isolates the neural network forward pass, the backend pose solver (for SCR methods), and the total end-to-end pipeline latency including data transfers and tensor preparation. All are median readings.}
\label{tab:sota_efficiency}
\small
\begin{tabular}{|l|c|c|c|c|}
\hline
\textbf{Method (A/S)} & \textbf{Network Pass (ms)} & \textbf{Backend Solver (ms)} & \textbf{End-to-End (ms)} & \textbf{Params (M)} \\
\hline\hline
DiffLoc~\cite{li2024diffloc} (A) & 53.87 & N/A & 54.05 & 40 \\
LiSA~\cite{10654844} (S)    & 32.90 & 5.40 & 41.30 & 105 \\
SGLoc~\cite{li2023sgloc} (S)   & 48.48 & 86.98 & 146 & 105 \\
LightLoc~\cite{li2025lightloc} (S)& 10.69 & 6.18 & 21.79 & 22 \\
HypLiLoc~\cite{wang2023hypliloc} (A)& 11.04 & N/A & \textbf{11.40} & 52 \\
\hline
MB-Loc (Ours) (S) & \textbf{6.80} & \textbf{1.80} & 21.13 & \textbf{16} \\
\hline
\end{tabular}
\end{center}
\end{table}
As reported in Table~\ref{tab:sota_efficiency}, MB-Loc completes its network forward pass in just 6.8~ms. This drastically outperforms competing 3D-SCR architectures like SGLoc~\cite{li2023sgloc} (48.48~ms) and LiSA~\cite{10654844} (32.9~ms), while also edging out the fastest APR baseline, HypLiLoc~\cite{wang2023hypliloc} (11.04 ms), and remaining highly competitive with lightweight SCR variants like LightLoc~\cite{li2025lightloc} (10.69 ms). Furthermore, our RANSAC backend resolves the global pose in a mere 1.8~ms. Ultimately, MB-Loc registers a total end-to-end pipeline latency of 21.1~ms, achieving a $2.6\times$ speedup over DiffLoc~\cite{li2024diffloc} and a $6.9\times$ speedup over SGLoc~\cite{li2023sgloc}.
MB-Loc achieves this real-time speed while maintaining a highly compact footprint. At just 16M parameters, it requires $6.5\times$ fewer parameters than SGLoc~\cite{li2023sgloc} \& LiSA~\cite{10654844} (105M) and $2.5\times$ fewer than DiffLoc~\cite{li2024diffloc} (40M), successfully breaking the computational limits that typically stall dense 3D localization pipelines.
\subsection{Ablation Study}
\label{subsec:ablation}

In this section, we conduct extensive ablation experiments to validate our architectural design choices and evaluate system sensitivity.
We initially analyse the trade-off between localization precision, runtime latency, and spatial compression by varying the number of Z-slicing planes $P \in \{5, 8, 10, 15\}$ and the horizontal grid resolution $G \in \{256, 512\}$ where $N_{\max} = 2,000$ \& $\tau=4.0$ m were used for these ablative experiments. 
\begin{table}[h]
\centering
\caption{Structural ablation where the \% of spatial data loss is computed relative to an average raw scan depth of 47,972 points. All time readings are median values.}
\label{tab:structural_ablation}
\resizebox{\linewidth}{!}{%
\begin{tabular}{|l|c|c|c|c|c|}
\hline
\textbf{Configuration} & \textbf{Mean Errors} & \textbf{Network Pass + RANSAC} & \textbf{End-to-End} & \textbf{Mean Total No. of} & \textbf{Spatial Data} \\
\textbf{$(P, G, G)$} & \textbf{(Trans. [m] / Rot. [$^{\circ}$])} & \textbf{(ms)} & \textbf{(ms)} & \textbf{Points Projected} & \textbf{Loss (\%)} \\
\hline\hline
$10 \times 256 \times 256$ & 1.49 / 2.57 & 6.47 & 8.59  & 7,799  & 83.74\% \\
$15 \times 256 \times 256$ & 1.11 / 2.31 & 6.90 & 8.76  & 8,554  & 82.17\% \\
$5 \times 512 \times 512$  & 1.12 / 2.42 & 8.35 & 11.68 & 12,692 & 73.54\% \\
$8 \times 512 \times 512$  & 1.08 / 2.44 & 8.40 & 13.31 & 13,552 & 71.75\% \\
$10 \times 512 \times 512$ & 1.02 / 2.28 & 8.48 & 14.48 & 13,998 & 70.82\% \\
$15 \times 512 \times 512$ & 0.91 / 2.30 & 8.60 & 21.13 & 14,870 & 69.00\% \\
\hline
\end{tabular}%
}
\end{table}

The empirical trends in Table~\ref{tab:structural_ablation} confirm that increasing the horizontal grid resolution 
from $G = 256$ to $G = 512$ yields a consistent improvement in localization 
accuracy which stems from a reduction in grid cell quantization 
collisions, where a finer grid better preserves subtle structural profiles that 
would otherwise be overwritten during downsampling. Concurrently, increasing the number of Z-slicing planes reduces translation errors across both grid resolutions. Splitting the $z$-axis into finer Z-slicing planes limits depth ambiguity within individual layers, providing more precise target labels for the coordinate regression task. Notably, even at the highest tested complexity ($15 \times 512 \times 512$) where spatial data retention is maximized (69\% loss), the complete computational core, from the 2D CNN forward pass to the closed-form pose resolution, executes in just 8.60~ms, demonstrating that the projection pipeline scales efficiently across all ablated configurations. We surpass the leading baseline architectures in both $10 \times 512 \times 512$ and $15 \times 512 \times 512$ configurations.

To assess how the hyperparameters inside our registration framework affect final pose estimation, we perform two isolated sensitivity tests: (1) varying the spatial inlier distance threshold $\tau$ while keeping the maximum correspondence subset size constant, and (2) varying the maximum correspondence subset size ($N_{\max}$) under a fixed inlier threshold. These evaluations are documented in Table~\ref{tab:ransac_threshold_ablation} and Table~\ref{tab:ransac_points_ablation} and have been performed on the $P=10$, $G=512$ configuration.

\begin{table}[h]
\centering
\begin{minipage}{0.48\linewidth}
\centering
\caption{RANSAC threshold sensitivity under a fixed point budget ($N_{\max} = 2,000$).}
\label{tab:ransac_threshold_ablation}
\small
\begin{tabular}{|c|c|c|}
\hline
\textbf{Threshold $\tau$ (m)} & \textbf{Mean Errors (m / $^{\circ}$)} \\
\hline\hline
1 & 1.07 / 2.53 \\
2.5 & 1.02 / 2.32\\
\textbf{4.0} & \textbf{1.02 / 2.28}\\
5.5 & 1.04 / 2.30\\
\hline
\end{tabular}
\end{minipage}
\hfill
\begin{minipage}{0.48\linewidth}
\centering
\caption{Max. correspondence sensitivity under a fixed inlier threshold ($\tau = 4.0$ m).}
\label{tab:ransac_points_ablation}
\small
\begin{tabular}{|c|c|c|}
\hline
\textbf{Max Corresp. $N_{\max}$} & \textbf{Mean Errors (m / $^{\circ}$)} \\ 
\hline\hline
100 & 1.04 / 2.36\\
\textbf{2,000} & \textbf{1.02 / 2.28}\\
6,000 & 1.1 / 2.37 \\
All Points & 1.14 / 2.4 \\
\hline
\end{tabular}
\end{minipage}
\end{table}
Ablative results on threshold sensitivity are reported in Table~\ref{tab:ransac_threshold_ablation}. A restrictive threshold of $\tau=1.0$~m degrades performance to 1.07~m, as it prematurely rejects valid correspondences from sparsely populated upper planes that exhibit natural coordinate variance due to grid quantization. 
Conversely, an overly permissive threshold of $\tau=5.5$~m offers no 
meaningful improvement over $\tau=4.0$~m while retaining more outliers. 
The optimal $\tau=4.0$~m strikes the critical balance between rejecting gross matching outliers while retaining a geometrically diverse inlier pool 
across all planes, allowing 
the backend to recover sub-meter global poses from individually noisy 
per-point predictions.

Table~\ref{tab:ransac_points_ablation} reports ablative results on the maximum correspondence subset size sensitivity. The performance peaks at $N_{\max}=2{,}000$ correspondences, with both 
sparser budgets ($N_{\max}=100$: 1.04~m) and larger budgets 
($N_{\max}=6{,}000$: 1.1~m) yielding degraded accuracy. Using all available valid correspondences without a budget cap degrades accuracy to 1.14~m, suggesting that an unbounded correspondence set introduces geometrically redundant and weakly-constrained point pairs 
that destabilize the SVD solution.

\section{Limitations and Future Work.} 
Our method relies on a fixed horizontal Z-axis slicing mechanism, implicitly 
assuming relatively planar terrain; in highly non-planar or off-road 
environments, fixed Z-slicing may fail to capture continuous geometric 
surfaces across elevation changes. As our method does not exploit multiple temporal frames at a time, severe dynamic occlusion is hard to handle. Additionally, when operating on 
pre-subsampled point-clouds, grid quantization collisions further 
compound sparsity, potentially reducing valid correspondences below 
the threshold required for robust RANSAC convergence. Finally, the 
fixed grid resolution $G$ imposes a hard quantization floor on metric 
precision that is irrecoverable by the SVD-RANSAC backend. Future work 
will explore adaptive density-driven slicing, temporal fusion across 
consecutive scans to mitigate dynamic occlusion, and dynamic grid 
resolution scaling based on local point-cloud density.
\section{Conclusion}
\label{sec:conclusion}
We presented MB-Loc, a lightweight Scene Coordinate Regression framework 
that achieves state-of-the-art outdoor LiDAR localization by operating 
entirely within the computational envelope of 2D convolutions. The 
combination of Z-axis slicing, 2.5D Multi-planar BEV projections, and deterministic latent regularization yields a system that 
is simultaneously geometrically faithful, robust to quantization noise 
and deployable in real-time. Extensive experiments on the NCLT dataset 
confirm that MB-Loc successfully breaks the accuracy-efficiency tradeoff
that has historically constrained dense 3D localization pipelines.

\appendix
\section*{Appendix}

\begin{figure}[h!]
\centering
   \includegraphics[width=1\linewidth]{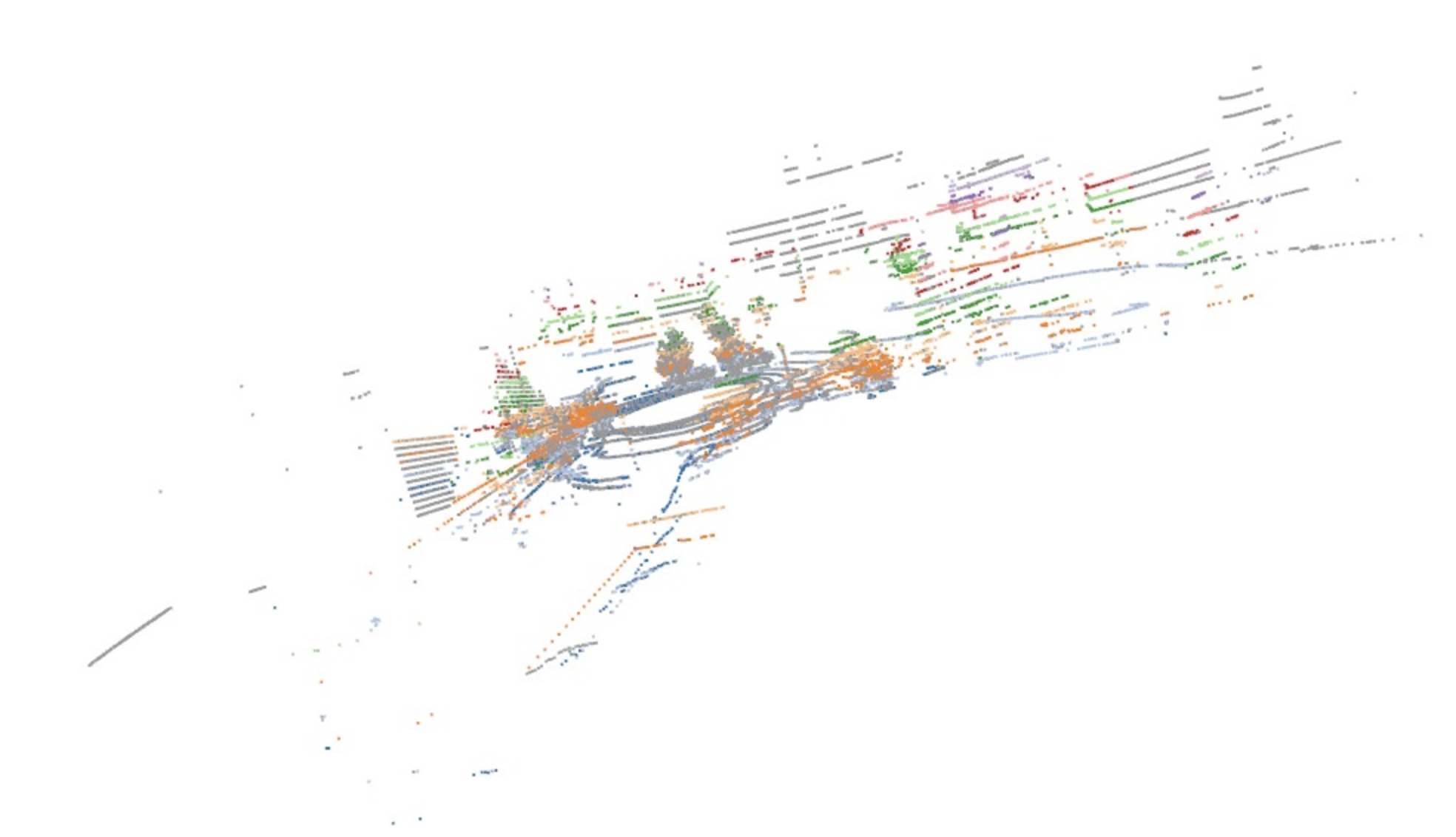}
    
    \vspace{0.5cm} 
    
    \includegraphics[width=1\linewidth]{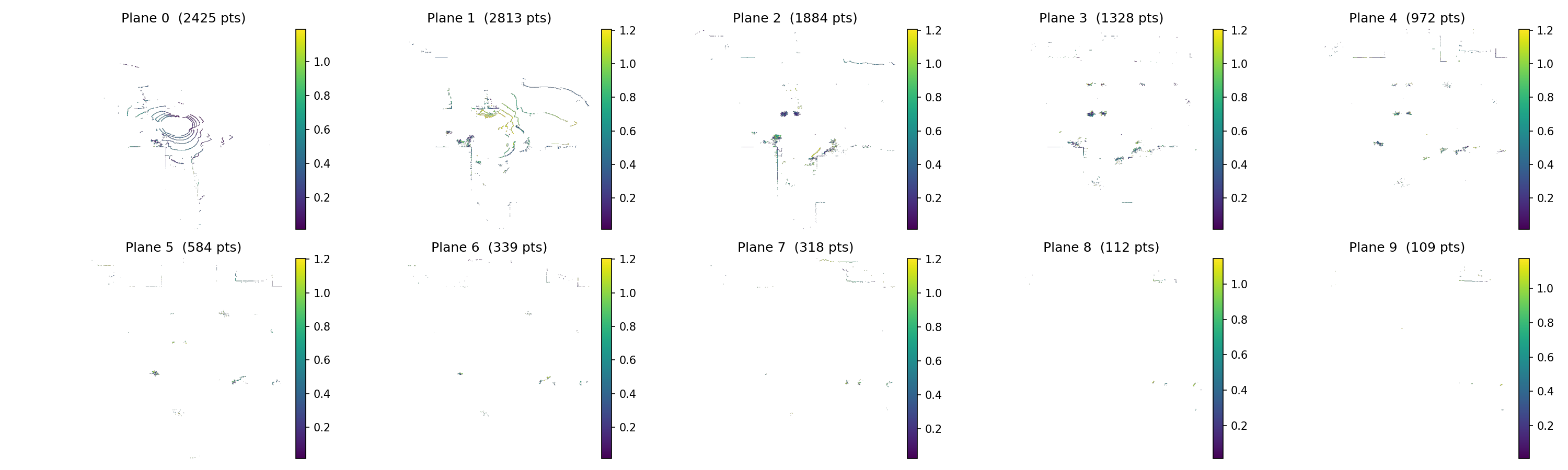}
    
\caption{Visualization of the Projection strategy. Top: Stratified 3D point cloud partitioned along the vertical Z-axis for a single scan. Bottom: Corresponding 2D multi-channel BEV grid projections across $P=10$ and $G=512$ on the NCLT~\cite{article} dataset.}
\label{fig:bev_visual}
\end{figure}

\section{Extended Details on the proposed MB-Loc Method} 
\label{sec:appendix_implementation}

Please note that the NCLT dataset~\cite{article} uses a \textit{North-East-Down} (NED) reference frame with the Velodyne HDL-32E mounted invertedly, conflicting with the standard $z$-up convention. We therefore apply $z' = -z$ to all raw coordinates prior to processing, ensuring that Z-axis slicing from $z_{max}$ downward and the subsequent 2D grid mapping reflect the true physical scene structure.

\subsection{Network Architecture}
\label{subsec:architecture_appendix}
Our proposed network utilizes a fully-convolutional encoder-decoder topology specifically optimized to translate multi-planar structural depth representations into dense coordinate offsets.
\subsubsection{Spatial Feature Encoder}
The encoder network sequentially downsamples the structured 2.5D input tensor $\mathcal{V} \in \mathbb{R}^{P \times G \times G}$ to extract rich geometric representations. The input first passes through an initial stem layer comprising a $3\times 3$ convolution with a stride of 2, projecting the $P$ input channels to a feature depth of 32. This is immediately followed by a 2D Batch Normalization layer, a LeakyReLU activation function, and a $3\times 3$ Max Pooling layer with a stride of 2.
The spatial dimensions are subsequently reduced by a further factor of $8\times$ through four consecutive stages of Residual blocks (the first stage operating at stride $1$ and the remaining three at stride $2$), yielding a cumulative downsampling of $32\times$ relative to the input. These blocks progressively expand the feature representation from 32 to 512 channels. To combat the structural sparsity inherent to outdoor LiDAR point-clouds, we integrate a Convolutional Block Attention Module (CBAM)~\cite{woo2018cbamconvolutionalblockattention} immediately following each of the four residual stages. It acts as a critical geometric gatekeeper: the channel attention sub-module isolates the most distinct horizontal Z-axis plane, while the spatial attention sub-module forces the network to focus on high-fidelity structural features while completely ignoring empty pixels. Given an intermediate feature map $\mathbf{F} \in \mathbb{R}^{C \times H' \times W'}$, the sequential attention alignment is mathematically formulated as:
\begin{equation}
    \mathbf{F}' = \mathbf{M}_c(\mathbf{F}) \otimes \mathbf{F}
    \label{eq:cbam_channel}
\end{equation}
\begin{equation}
    \mathbf{F}'' = \mathbf{M}_s(\mathbf{F}') \otimes \mathbf{F}'
    \label{eq:cbam_spatial}
\end{equation}
where $\otimes$ denotes element-wise multiplication. The channel attention map $\mathbf{M}_c \in \mathbb{R}^{C \times 1 \times 1}$ and spatial attention map $\mathbf{M}_s \in \mathbb{R}^{1 \times H' \times W'}$ are computed via:
\begin{equation}
    \mathbf{M}_c(\mathbf{F}) = \sigma\left( \text{MLP}(\text{AvgPool}(\mathbf{F})) + \text{MLP}(\text{MaxPool}(\mathbf{F})) \right)
    \label{eq:cbam_mc}
\end{equation}
\begin{equation}
    \mathbf{M}_s(\mathbf{F}') = \sigma\left( f^{7\times 7}\left([\text{AvgPool}(\mathbf{F}'); \text{MaxPool}(\mathbf{F}')]\right) \right)
    \label{eq:cbam_ms}
\end{equation}
where $\sigma$ represents the standard sigmoid activation function, $[\cdot ; \cdot]$ denotes a channel-wise concatenation operator, and $f^{7\times 7}$ signifies a 2D convolutional operation utilizing a $7\times 7$ kernel. The final output of the encoder is a highly condensed latent geometric map $\mathcal{F} \in \mathbb{R}^{512 \times \frac{G}{32} \times \frac{G}{32}}$.
\subsubsection{Regularized Deterministic Latent Bottleneck}
To prevent the network from over-fitting to transient noise or sparse dynamic artifacts, we pass $\mathcal{F}$ through a specialized latent bottleneck. Rather than executing a direct deterministic mapping or injecting stochastic variation via a probabilistic reparameterization vector, our bottleneck models structural uncertainty deterministically. We employ three parallel two-layer $1 \times 1$ convolutional projection heads (MLPs) to map $\mathcal{F}$ to a tightly bounded latent representation. Each head comprises a $1 \times 1$ convolution, a 2D Batch Normalization layer, and a LeakyReLU activation, followed by a second $1 \times 1$ convolution; the heads preserve the input dimensions, yielding tensors $\boldsymbol{\mu}, \boldsymbol{\sigma}, \mathbf{s} \in \mathbb{R}^{512 \times \frac{G}{32} \times \frac{G}{32}}$:
\begin{align}
    \boldsymbol{\mu} &= \text{MLP}_{\mu}(\mathcal{F}) \label{eq:bottleneck_mu}\\
    \boldsymbol{\sigma} &= \text{Softplus}\left(\text{MLP}_{\sigma}(\mathcal{F})\right) \label{eq:bottleneck_sigma}\\
    \mathbf{s} &= \text{Clamp}\left(\text{ReLU}\left(\text{MLP}_{s}(\mathcal{F})\right), 0, s_{max}\right) \label{eq:bottleneck_s}
\end{align}
The regularized latent state $\mathcal{Z} \in \mathbb{R}^{512 \times \frac{G}{32} \times \frac{G}{32}}$ passed to the decoder is then constructed through a deterministic scaling function:
\begin{equation}
    \mathcal{Z} = \boldsymbol{\mu} + (\mathbf{s} \odot \boldsymbol{\sigma})
    \label{eq:bottleneck_z}
\end{equation}
where $\odot$ represents the Hadamard product. While the network could trivially minimize $L_{coord}$ by collapsing $\boldsymbol{\sigma}$ toward zero and encoding all structure in $\boldsymbol{\mu}$, the KL penalty explicitly prevents this by penalizing deviations of $\boldsymbol{\sigma}$ from a unit normal prior, preserving the uncertainty-aware structure of the latent space. While the Deterministic Information Bottleneck~\cite{alemi2019deepvariationalinformationbottleneck,tishby2015deep} and disentangled VAE variants such as DIP-VAE~\cite{kumar2017variational} also regularize latent representations without full generative objectives, they either require a discrete compression target or impose second-moment matching constraints that are ill-suited to dense metric regression. Stochastic reparameterization, as utilized in standard VAEs~\cite{kingma2013auto}, introduces sampling variance during training that destabilizes the per-point coordinate  offset predictions, and while inference typically uses $\boldsymbol{\mu}$ directly, the stochastic training signal still induces suboptimal convergence for dense metric regression targets. By contrast, our deterministic bottleneck retains the KL penalty purely as a feature-smoothing regularizer, enforcing a smooth, continuous spatial manifold over the latent space without injecting any stochastic noise into the inference path, making it directly compatible with the closed-form pose solver downstream. Consequently, by enforcing a Softplus activation on the spatial variance proxy $\boldsymbol{\sigma}$ and applying a hard upper-bound saturation threshold $s_{max}$ via a clamped ReLU on the learned scaling factor $\mathbf{s}$, the bottleneck effectively dampens the influence of poorly-constrained structural regions (such as open air or reflective surfaces) without introducing stochastic instability into the localization pipeline.
\subsubsection{Coordinate Regression Decoder}
The decoder network takes the regularized latent representation $\mathcal{Z}$ and expands it back to the native spatial resolution of the input grid. This spatial inflation is achieved through a sequence of five $4 \times 4$ transposed convolutional layers, each configured with a stride of 2 and padding of 1 to sequentially double the spatial dimensions ($32\times$ total upsampling). Every transposed convolutional layer is followed by a 2D Batch Normalization layer and a LeakyReLU activation function.
A terminal $1 \times 1$ convolutional layer maps the upsampled features directly to the target regression dimensions without an activation function, yielding the final dense prediction tensor $\hat{\Delta} \in \mathbb{R}^{3P \times G \times G}$. This tensor is subsequently reshaped to a dimension of $\mathbb{R}^{P \times 3 \times G \times G}$, mapping directly to the individual horizontal Z-slicing planes:
\begin{equation}
    \hat{\Delta}(k, \cdot, u, v) = \begin{bmatrix} \hat{\Delta} x \\ \hat{\Delta} y \\ \hat{\Delta} z \end{bmatrix}
    \label{eq:decoder_output}
\end{equation}
This output provides a dense, per-point prediction of the 3D spatial offsets required to map the local coordinate tensor $\mathcal{C}_{LiDAR}$ back into the global world coordinates.

\subsection{Pose Optimization using RANSAC}
\label{subsec:pose_appendix}
Once the network infers the dense spatial offsets $\hat{\Delta}$, estimating the global 6-DoF pose becomes a 3D-to-3D  correspondence registration problem. We utilize the validity mask $\mathcal{M}$ to extract the occupied physical points from the local coordinate tensor $\mathcal{C}_{LiDAR}$, yielding a set of source points $\mathcal{C}_{LiDAR}$. Concurrently, we construct the predicted global target points as $\hat{\mathcal{C}}_{world} = \mathcal{C}_{LiDAR} + \hat{\Delta}$.  Because the network predicts offsets per grid cell, the source and target point sets maintain a strict one-to-one correspondence for all retained structural points.
To compute the rigid transformation matrix from these correspondences while aggressively filtering out regression outliers and quantization noise, we employ a RANSAC-based~\cite{10.1145/358669.358692} estimator. In each iteration, we sample a minimal set of $N=3$ point correspondences to compute a candidate rigid transformation via closed-form SVD. First, we compute the centroids of the sampled source and target points:
\begin{equation}
    \bar{\mathbf{c}}_{LiDAR} = \frac{1}{N} \sum_{i=1}^N \mathbf{c}_{LiDAR}^{(i)}, \quad \bar{\mathbf{c}}_{world} = \frac{1}{N} \sum_{i=1}^N \hat{\mathbf{c}}_{world}^{(i)}
    \label{eq:ransac_centroids}
\end{equation}
Next, we construct the cross-covariance matrix $\mathbf{H} \in \mathbb{R}^{3 \times 3}$:
\begin{equation}
    \mathbf{H} = \sum_{i=1}^N \left(\mathbf{c}_{LiDAR}^{(i)} - \bar{\mathbf{c}}_{LiDAR}\right) \left(\hat{\mathbf{c}}_{world}^{(i)} - \bar{\mathbf{c}}_{world}\right)^T
    \label{eq:ransac_H}
\end{equation}
We decompose the covariance matrix using SVD such that $\mathbf{H} = \mathbf{U} \boldsymbol{\Sigma} \mathbf{V}^T$. The optimal rotation matrix $\mathbf{R}$ is extracted utilizing Kabsch's algorithm~\cite{1976AcCrA..32..922K}:
\begin{equation}
    \mathbf{R} = \mathbf{V} \begin{pmatrix} 1 & 0 & 0 \\ 0 & 1 & 0 \\ 0 & 0 & \det(\mathbf{V}\mathbf{U}^T) \end{pmatrix} \mathbf{U}^T
    \label{eq:ransac_R}
\end{equation}
The translation vector $\mathbf{t}$ is then straightforwardly derived from the aligned centroids:
\begin{equation}
    \mathbf{t} = \bar{\mathbf{c}}_{world} - \mathbf{R} \bar{\mathbf{c}}_{LiDAR}
    \label{eq:ransac_t}
\end{equation}
To maximize computational efficiency, we employ standard early termination logic. The minimum required iterations $k$ is updated dynamically to guarantee the desired confidence level, which is given by: 
\begin{equation}
    k = \frac{\log(1-p)}{\log(1-w^3)}
\end{equation}

where $p$ is the target confidence probability and $w$ is the dynamically estimated inlier ratio based on a predefined distance threshold.

\setcounter{section}{0}
\setcounter{table}{0}
\setcounter{figure}{0}
\setcounter{equation}{0}
\renewcommand{\thesection}{S\arabic{section}}
\renewcommand{\thesubsection}{S\arabic{section}.\arabic{subsection}}
\renewcommand{\thetable}{S\arabic{table}}
\renewcommand{\thefigure}{S\arabic{figure}}
\renewcommand{\theequation}{S\arabic{equation}}

\renewcommand{\theHsection}{supp.\arabic{section}}
\renewcommand{\theHsubsection}{supp.\arabic{section}.\arabic{subsection}}
\renewcommand{\theHtable}{supp.table.\arabic{table}}
\renewcommand{\theHfigure}{supp.figure.\arabic{figure}}
\renewcommand{\theHequation}{supp.equation.\arabic{equation}}

\section*{Supplementary Material}

This supplementary material complements the main paper with additional context. We
provide a consolidated, layer-wise view of the network
(Section~\ref{sec:supp_architecture}), collect additional implementation, training, and
evaluation details (Section~\ref{sec:supp_additional}), expand on the NCLT~\cite{article} benchmark dataset, preprocessing pipeline, and per-sequence results
(Section~\ref{sec:supp_dataset}), and report an evaluation on the
subsampled Oxford Radar RobotCar~\cite{barnes2020oxford} dataset (Section~\ref{sec:supp_oxford}). Our codebase will be made publicly available upon the publication of this work.

\section{Network Architecture}
\label{sec:supp_architecture}

MB-Loc is a fully-convolutional (ResNet) encoder--decoder that maps the multi-planar structural
depth representation to dense per-point 3D coordinate offsets. The formal definitions of
the attention operator, the latent bottleneck, and the decoder head are provided in the
Network Architecture subsection of the main paper's appendix. A consolidated layer-wise summary is given in Table~\ref{tab:supp_arch}.

\begin{table}[h]
\centering
\caption{Layer-wise summary of the MB-Loc network. $P$ is the number of
Z-slicing planes and $G$ the horizontal grid resolution; spatial sizes are given for a
$G \times G$ input grid.}
\label{tab:supp_arch}
\small
\begin{tabular}{|l|c|c|}
\hline
\textbf{Stage} & \textbf{Output Channels} & \textbf{Spatial Resolution} \\
\hline\hline
Input                  & $P$  & $G \times G$ \\
Stem (conv + max-pool) & 32   & $G/4 \times G/4$ \\
\hline
Encoder Stage 1        & 64   & $G/4 \times G/4$ \\
Encoder Stage 2        & 128  & $G/8 \times G/8$ \\
Encoder Stage 3        & 256  & $G/16 \times G/16$ \\
Encoder Stage 4        & 512  & $G/32 \times G/32$ \\
\hline
Latent bottleneck      & 512  & $G/32 \times G/32$ \\
\hline
Decoder                & 32   & $G \times G$ \\
Head ($1\times1$)      & $3P$ & $G \times G$ \\
\hline
\end{tabular}
\end{table}

\section{Additional Details}
\label{sec:supp_additional}

\subsection{Implementation and Hyperparameters}
\label{subsec:supp_impl}

The hyperparameters chosen for our optimal configuration are shown in Table~\ref{tab:supp_hparams}.

\begin{table}[h]
\centering
\caption{Default hyperparameter configuration for MB-Loc on the NCLT~\cite{article}
dataset.}
\label{tab:supp_hparams}
\small
\begin{tabular}{|l|c|}
\hline
\textbf{Hyperparameter} & \textbf{Value} \\
\hline\hline
Z-slicing planes $P$               & 15 \\
Horizontal grid resolution $G$     & 512 \\
Optimizer                          & Adam \\
Initial learning rate              & $3 \times 10^{-3}$ \\
Weight decay                       & $1 \times 10^{-6}$ \\
LR scheduler                       & Step ($\times 0.85$ every 20 epochs) \\
KL regularization weight $\lambda$ & $1 \times 10^{-4}$ \\
RANSAC inlier threshold $\tau$     & $4.0$\,m \\
Max. correspondences $N_{\max}$    & 2000 \\
RANSAC sample size $N$             & 3 \\
RANSAC target confidence $p$       & 0.95 \\
\hline
\end{tabular}
\end{table}

\subsection{Spatial Augmentation Strategy}
\label{subsec:supp_aug}

The spatial augmentations detailed in the main paper are applied prior to planar
projection. Because the augmentation precedes projection, the regression target (the
offsets) stays geometrically consistent with the augmented input, so the network is
supervised to localize regardless of the sensor's instantaneous viewpoint.

\subsection{Training Objective and Pose Solver}
\label{subsec:supp_objective}

The complete training objective (the mask-gated L1 coordinate loss and the KL
regularizer balanced by $\lambda$) and the step-by-step pose-solver derivation
(centroids, cross-covariance, rotation/translation recovery, and the iteration-count
update) are provided in the main paper and its appendix.

\section{NCLT Dataset and Evaluation}
\label{sec:supp_dataset}

We evaluate the proposed MB-Loc for LiDAR localization on large-scale outdoor benchmark datasets: the NCLT~\cite{article} and Oxford Radar RobotCar~\cite{barnes2020oxford} datasets. Collected on the University of Michigan's North Campus, the NCLT dataset is recorded by
sensors mounted on a Segway robotic platform. It comprises 27 traversals, each spanning
an area of $0.45$\,km$^2$, with the point-clouds captured by a single Velodyne HDL-32E
LiDAR.

\textbf{NCLT Train/Test Split.}
The NCLT train/test split is summarized in Table~\ref{tab:supp_nclt}. The 2012-05-26
session is the most temporally distant from training and is correspondingly the hardest,
which is why the competing baselines in the main paper degrade most sharply on it.

\begin{table}[h]
\centering
\caption{Dataset details on the NCLT~\cite{article} dataset.}
\label{tab:supp_nclt}
\small
\begin{tabular}{|l|c|c|c|c|}
\hline
\textbf{Sequence} & \textbf{Length} & \textbf{Tag} & \textbf{Training} & \textbf{Test} \\
\hline\hline
2012-01-22 & 6.1\,km & overcast & \checkmark & \\
2012-02-02 & 6.2\,km & sunny    & \checkmark & \\
2012-02-18 & 6.2\,km & sunny    & \checkmark & \\
2012-05-11 & 6.0\,km & sunny    & \checkmark & \\
\hline
2012-02-12 & 5.8\,km & sunny    & & \checkmark \\
2012-02-19 & 6.2\,km & overcast & & \checkmark \\
2012-03-31 & 6.0\,km & overcast & & \checkmark \\
2012-05-26 & 6.3\,km & sunny    & & \checkmark \\
\hline
\end{tabular}
\end{table}

\textbf{Coordinate Frame Handling.}
The reasoning behind our sign-inversion preprocessing step ($z' = -z$, which reconciles
the inverted NED mounting of the HDL-32E with our $z$-up Z-slicing convention) is given
in the Extended Details appendix of the main paper.

\textbf{Projection Statistics.}
As quantified in the structural ablation of the main paper, our intentionally lossy,
collision-aware projection discards roughly $69\%$--$84\%$ of the points in an average
raw NCLT scan (depending on the configuration $(P, G, G)$), while preserving the
structural detail that matters for localization.

\subsection{Per-Sequence Quantitative Results}
\label{subsec:supp_persequence}

To complement the averaged results in the main paper, Table~\ref{tab:supp_persequence}
provides a per-sequence breakdown of MB-Loc across the structural configurations
$(P, G, G)$, on each of the four NCLT~\cite{article} evaluation sessions. Accuracy improves consistently with finer grids and more Z-slicing planes, and our default $15\times512\times512$ configuration attains the best
average error.
\begin{table}[h]
\centering
\caption{Per-sequence localization accuracy of MB-Loc on the NCLT~\cite{article}
evaluation sessions across structural configurations $(P, G, G)$. Each entry is the mean
translation\,/\,rotation error in m\,/\,$^{\circ}$; the last column is the average over
the four sessions. RANSAC settings: $\tau = 4.0$\,m, $N_{\max} = 2{,}000$, $p = 0.95$.
Our default configuration is in \textit{italics}.}
\label{tab:supp_persequence}
\small
\begin{tabular}{|l|c|c|c|c|c|}
\hline
\textbf{Config $(P,G,G)$} & \textbf{2012-02-12} & \textbf{2012-02-19} & \textbf{2012-03-31} & \textbf{2012-05-26} & \textbf{Avg [m/$^{\circ}$]} \\
\hline\hline
$10\times256\times256$ & 1.50/2.70 & 1.38/2.50 & 1.46/2.50 & 1.65/2.60 & 1.49/2.57 \\
$15\times256\times256$ & 1.13/2.40 & 1.05/2.24 & 1.07/2.27 & 1.21/2.34 & 1.11/2.31 \\
$5\times512\times512$  & 1.16/2.51 & 1.12/2.34 & 1.08/2.51 & 1.15/2.34 & 1.12/2.42 \\
$8\times512\times512$  & 1.17/2.50 & 1.03/2.35 & 1.06/2.57 & 1.06/2.34 & 1.08/2.44 \\
$10\times512\times512$ & 1.08/2.38 & 1.01/2.23 & 0.99/2.32 & 1.03/2.21 & 1.02/2.28 \\
\hline
$\mathit{15\times512\times512}$ & \textit{0.96/2.37} & \textit{0.88/2.22} & \textit{0.89/2.40} & \textit{0.93/2.23} & \textit{0.91/2.30} \\
\hline
\end{tabular}
\end{table}

\subsection{Effect of the KL Regularizer}
\label{subsec:supp_kl}

To isolate the contribution of the bottleneck KL term, we retrain the
$10\times256\times256$ configuration on NCLT~\cite{article} with the KL regularizer
disabled ($\lambda = 0$) while holding all other settings fixed
(Table~\ref{tab:supp_kl}). Removing it degrades accuracy on every evaluation session,
and the loss is largest on the hardest, most temporally distant 2012-05-26 sequence.
Without the KL term the bottleneck variance is free to collapse and the latent manifold
over-fits the sparse, noisy returns, so the smoothing it provides yields a small but
consistent improvement in cross-session generalization, supporting its inclusion in our
default objective.

\begin{table}[h]
\centering
\caption{Effect of the KL regularizer on MB-Loc ($10\times256\times256$,
NCLT~\cite{article}).}
\label{tab:supp_kl}
\small
\begin{tabular}{|l|c|c|c|c|c|}
\hline
\textbf{Variant} & \textbf{2012-02-12} & \textbf{2012-02-19} & \textbf{2012-03-31} & \textbf{2012-05-26} & \textbf{Avg [m/$^{\circ}$]} \\
\hline\hline
With KL    & 1.50/2.70 & 1.38/2.50 & 1.46/2.50 & 1.65/2.60 & 1.49/2.57 \\
Without KL & 1.66/3.04 & 1.50/2.81 & 1.63/2.87 & 2.02/3.29 & 1.70/3.00 \\
\hline
\end{tabular}
\end{table}

\subsection{Reconstruction Fidelity of Predicted Correspondences}
\label{subsec:supp_fidelity}

\begin{figure}[h]
\centering
\includegraphics[width=\linewidth]{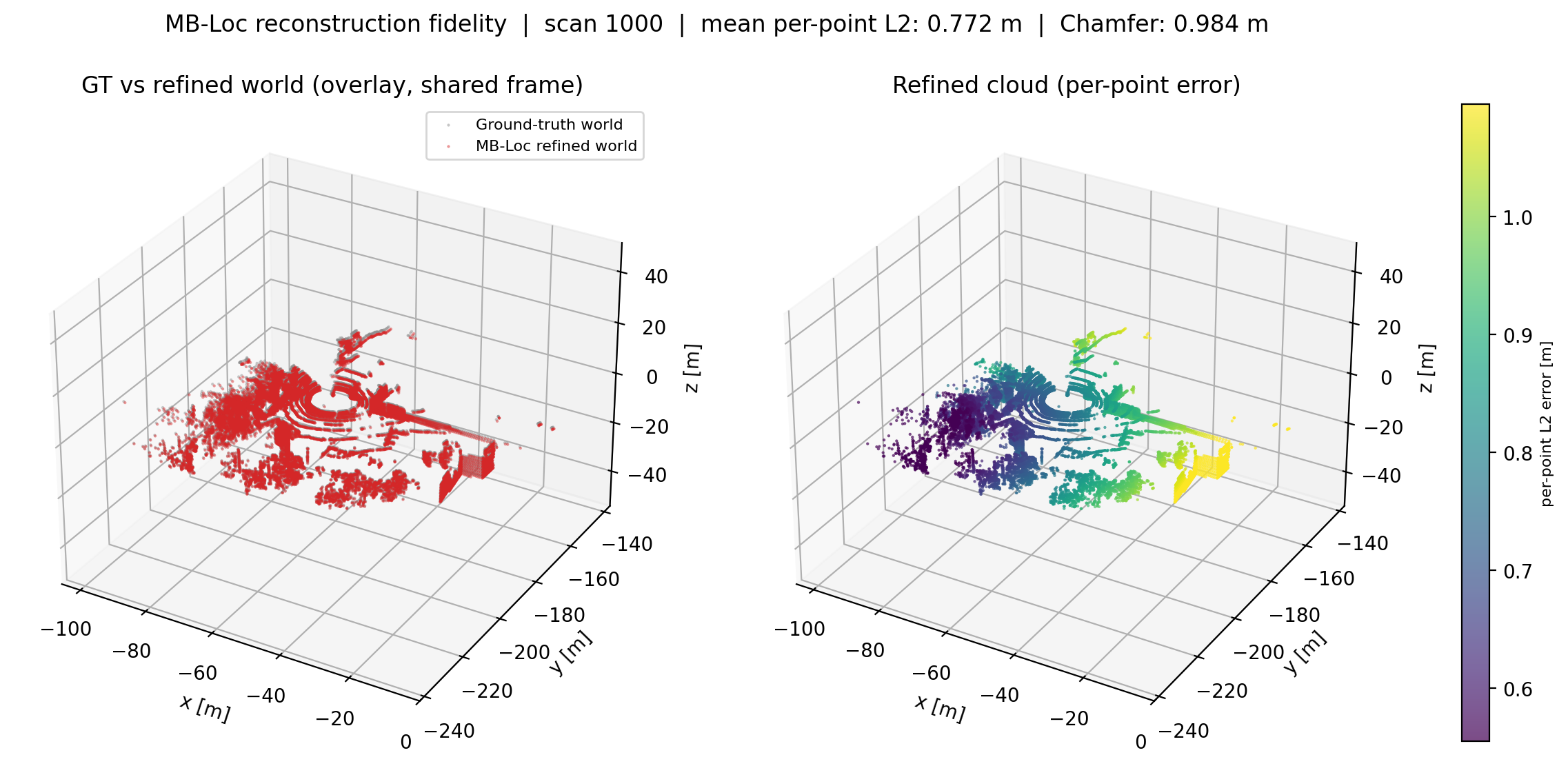}
\caption{Reconstruction fidelity on a scan from the NCLT~\cite{article} 2012-02-12
sequence ($P=15$, $G=512$; RANSAC $\tau=4.0$\,m, $N_{\max}=2{,}000$, $p=0.95$).
\textbf{Left:} ground-truth world-cloud (gray) overlaid with the MB-Loc RANSAC-refined-cloud (red) in a shared world frame. \textbf{Right:} the same refined cloud colored by
per-point $L_2$ error; the error grows with range owing to the rotational lever-arm. Both clouds are the valid
projected points retained after Z-slicing and scatter-reduce (not the raw input scan).
The annotated per-scan values are $0.77$\,m mean per-point $L_2$ and $0.98$\,m Chamfer.}
\label{fig:supp_fidelity}
\end{figure}

To verify that MB-Loc's dense predicted correspondences reconstruct the true scene
geometry, rather than merely yielding an accurate pose by chance, we directly assess
the quality of the regressed world coordinates before they are collapsed into a single
pose. Using our default configuration, we run the
full inference pipeline on the NCLT~\cite{article} 2012-02-12 sequence. For each scan, we recover the rigid transform with our pose estimation backend, apply it to the input LiDAR
points to obtain the refined world-cloud $\hat{\mathcal{C}}_{world}$, and compare it
against the ground-truth world-cloud $\mathcal{C}_{world}$ using two complementary
metrics: (i) the \emph{mean per-point $L_2$ error}, i.e. the average
$\lVert \hat{\mathcal{C}}_{world} - \mathcal{C}_{world}\rVert_2$ taken over the matched
(strictly one-to-one) correspondences; and (ii) the \emph{symmetric Chamfer distance}, a
nearest-neighbour shape-agreement measure that does not assume a known correspondence.
Aggregated over $2{,}001$ scans sampled uniformly at random across the sequence, MB-Loc
attains a mean per-point error of $1.18 \pm 0.98$\,m and a Chamfer distance of
$1.15 \pm 0.46$\,m. The visualization for a single scan is shown in Figure~\ref{fig:supp_fidelity}.

\textbf{Relationship to the pose error in the main paper.} These per-point and Chamfer
values are \emph{cloud-level reconstruction} metrics and should not be conflated with
the $6$-DoF pose error reported in the main paper, where MB-Loc achieves
$0.96$\,m\,/\,$2.37^{\circ}$ on this same 2012-02-12 sequence. The pose translation error
measures only the displacement of the single recovered sensor origin, whereas the
per-point $L_2$ measures the displacement of \emph{every} reconstructed point.

\subsection{Additional Qualitative Trajectories}
\label{subsec:supp_qualitative}

\begin{figure}[p]
\centering
\includegraphics[width=0.9\linewidth]{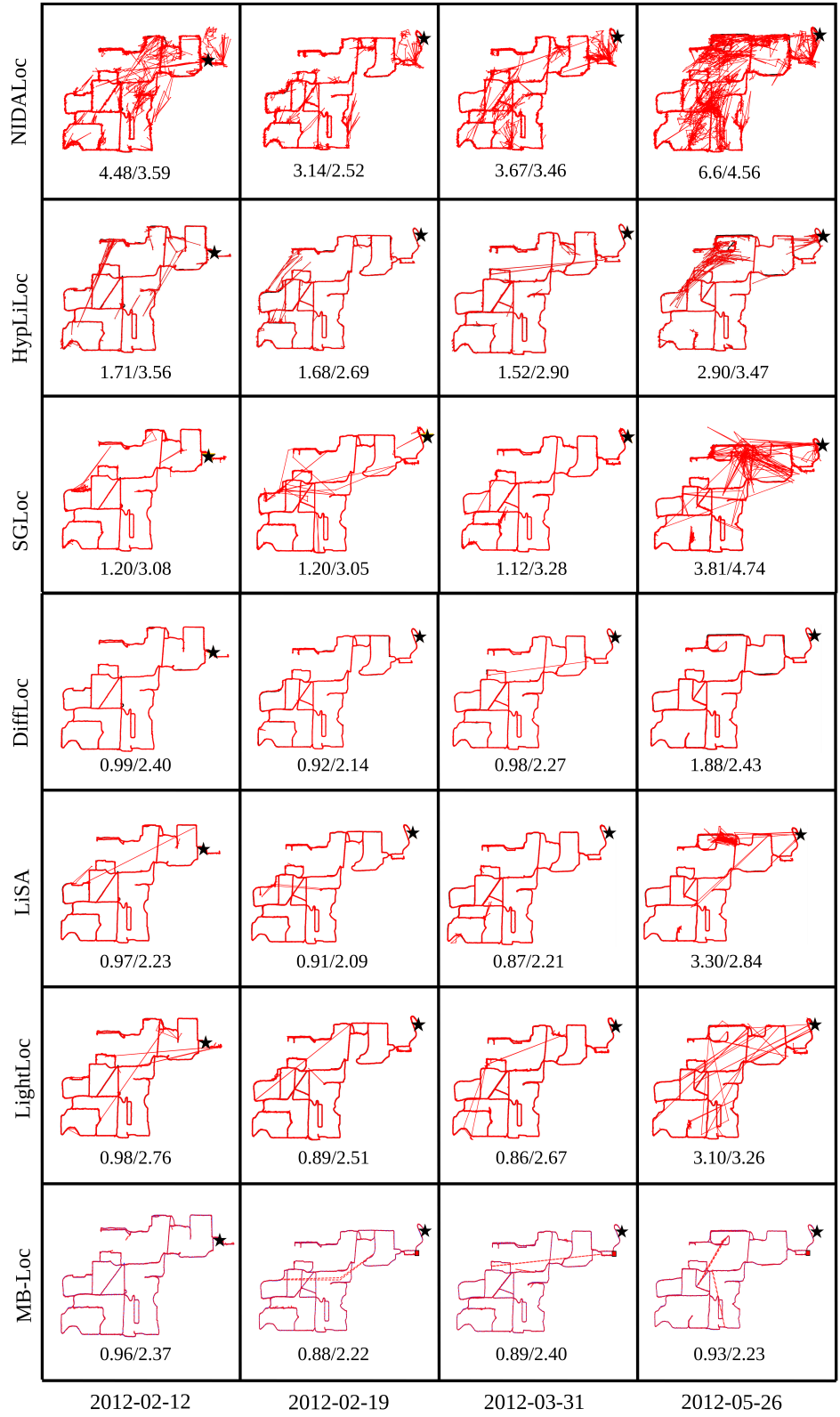}
\caption{Predicted (red) vs.\ ground-truth (black) trajectories on the four
NCLT evaluation sessions (columns) for the different baselines (rows). The mean translation\,/\,rotation error (m\,/\,$^{\circ}$) is annotated below each plot, and $\bigstar$ marks the first frame.}
\label{fig:supp_traj_a}
\end{figure}

Figure~\ref{fig:supp_traj_a} compares full predicted
trajectories against ground truth across all four NCLT~\cite{article} evaluation
sessions and against contemporary APR and SCR baselines. Across every session, including
the most challenging 2012-05-26 sequence, MB-Loc stays close to the ground-truth path,
whereas competing methods exhibit pronounced drift and large-magnitude outlier spikes.

\section{Oxford RobotCar (Subsampled) Dataset and Evaluation}
\label{sec:supp_oxford}

As noted in the main manuscript, access to the full, raw Oxford Radar RobotCar dataset~\cite{barnes2020oxford} was restricted prior to our initial submission, limiting our primary evaluation to the NCLT~\cite{article} benchmark. However, to ensure a rigorous and comprehensive comparison against recent state-of-the-art methods, we subsequently conducted additional evaluations using a heavily subsampled version of the Oxford dataset.

The Oxford Radar RobotCar dataset captures complex urban conditions with severe weather and traffic variations. The specific subsampled variant we utilize was officially processed and publicly released by the authors of HypLiLoc~\cite{wang2023hypliloc} on their GitHub. In this version, every LiDAR scan has been drastically downsampled to exactly 4,096 points. While this removes the vast majority of the dense geometric structure typically relied upon by SCR methods, it serves as an extreme stress test for the robustness of our 2.5D Multi-planar BEV representation under severe spatial sparsity.
We evaluated MB-Loc on this subsampled dataset using the train/test split summarized in Table~\ref{tab:supp_oxford}.
\begin{table}[h]
\centering
\caption{Dataset details on the Oxford~\cite{barnes2020oxford} dataset.}
\label{tab:supp_oxford}
\small
\begin{tabular}{|l|c|c|c|c|}
\hline
\textbf{Sequence} & \textbf{Length} & \textbf{Tag} & \textbf{Training} & \textbf{Test} \\
\hline\hline
11-14-02-26 & 9.37\,km & sunny    & \checkmark & \\
14-12-05-52 & 9.22\,km & overcast & \checkmark & \\
14-14-48-55 & 9.04\,km & overcast & \checkmark & \\
18-15-20-12 & 9.04\,km & overcast & \checkmark & \\
\hline
15-13-06-37 & 8.85\,km & overcast & & \checkmark \\
17-14-03-00 & 9.02\,km & sunny    & & \checkmark \\
18-14-14-42 & 9.04\,km & overcast & & \checkmark \\
\hline
\end{tabular}
\end{table}

\begin{figure}[p]
\centering
\includegraphics[width=0.8\linewidth]{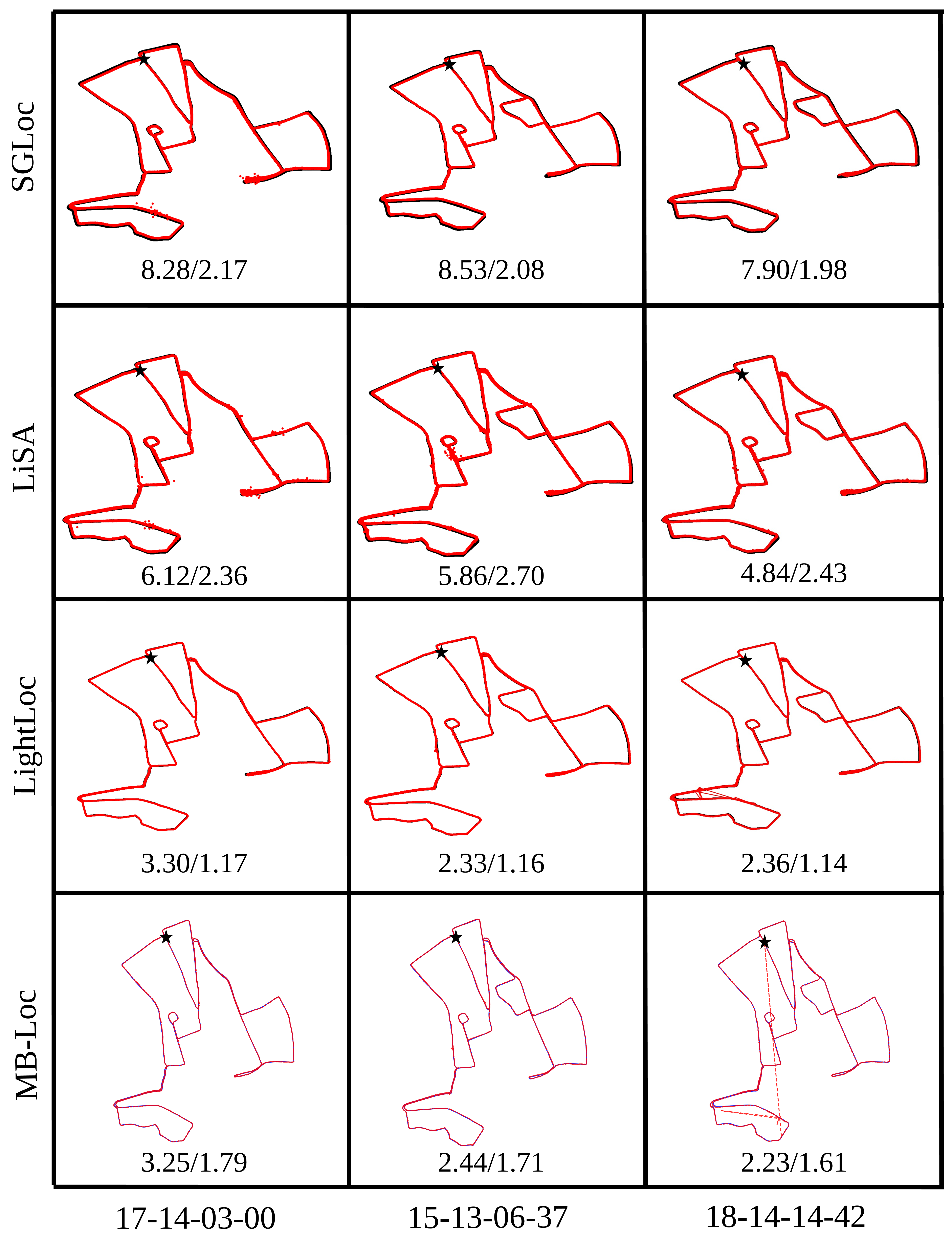}
\caption{Predicted (red) vs.\ ground-truth (black) trajectories on the subsampled Oxford dataset. The mean translation\,/\,rotation error (m\,/\,$^{\circ}$) is annotated below each plot, and $\bigstar$ marks the first frame.}
\label{fig:supp_traj_oxford}
\end{figure}

\textbf{Localization Results.}
Table~\ref{tab:supp_oxford_results} reports the per-sequence localization accuracy of
MB-Loc against recent state-of-the-art APR and SCR baselines on the three subsampled
Oxford~\cite{barnes2020oxford} test sequences. The dominant trend we observe is that
most state-of-the-art dense 3D architectures degrade sharply under this extreme
$4{,}096$-point subsampling: the dense geometric structure they depend on is largely
stripped away, and methods such as SGLoc~\cite{li2023sgloc} and LiSA~\cite{10654844} incur errors of several
metres. The clear exception is LightLoc~\cite{li2025lightloc}, which remains robust
under severe sparsity. MB-Loc is competitive with the other baselines but, like the other
dense-prediction methods, also degrades markedly relative to its NCLT performance and
does not match LightLoc here. This is consistent with the limitation noted in the main
paper that on pre-subsampled point-clouds, grid-quantization collisions further compound the
sparsity and can drive the number of valid correspondences below the threshold required
for robust RANSAC convergence. Qualitative predicted-versus-ground-truth trajectories for the dataset are shown in
Figure~\ref{fig:supp_traj_oxford}.

\begin{table}[h]
\centering
\caption{Localization accuracy on the subsampled Oxford~\cite{barnes2020oxford} dataset.
Each entry is the mean translation\,/\,rotation error in m\,/\,$^{\circ}$; the last column
is the average over the three test sequences. We highlight the
\textcolor{blue}{best} and \textcolor{magenta}{second best} results.}
\label{tab:supp_oxford_results}
\small
\begin{tabular}{|l|c|c|c|c|c|}
\hline
\textbf{Method} & \textbf{Mech.} & \textbf{17-14-03-00} & \textbf{15-13-06-37} & \textbf{18-14-14-42} & \textbf{Average [m/$^{\circ}$]} \\
\hline\hline
SGLoc     & S & 8.28/2.17  & 8.53/2.08  & 7.9/1.98   & 8.23/2.07 \\
LiSA      & S & 6.12/2.36  & 5.86/2.70  & 4.84/2.43  & 5.60/2.49 \\
LightLoc  & S & \textcolor{magenta}{3.30}/\textcolor{blue}{1.17}  & \textcolor{blue}{2.33}/\textcolor{blue}{1.16}  & \textcolor{magenta}{2.36}/\textcolor{blue}{1.14}  & \textcolor{magenta}{2.66}/\textcolor{blue}{1.15} \\
\hline
MB-Loc    & S & \textcolor{blue}{3.25}/\textcolor{magenta}{1.79}  & \textcolor{magenta}{2.44}/\textcolor{magenta}{1.71}  & \textcolor{blue}{2.23}/\textcolor{magenta}{1.61}  & \textcolor{blue}{2.64}/\textcolor{magenta}{1.70} \\
\hline
\end{tabular}
\end{table}

\textbf{Baseline and MB-Loc Implementation Details.}
We train the baselines from scratch on the identical subsampled splits ($4{,}096$ points per scan, stored as 3-channel $(x,y,z)$
\texttt{.npy} files) using their official public implementation. We train SGLoc~\cite{li2023sgloc} and LiSA~\cite{10654844} at a voxel size of $0.3$\,m. We train LightLoc~\cite{li2025lightloc} at a voxel size of $0.25$\,m; it freezes its scene-agnostic feature
backbone and trains only the scene-specific prediction heads, and applies its
redundant-sample downsampling (prune ratio $0.25$). DiffLoc~\cite{li2024diffloc} is the only APR baseline and
follows a markedly different pipeline. It converts each scan into a $32\times512$
spherical range image with five channels (range, $x$, $y$, $z$, intensity) and regresses
the pose using a DINOv2-pretrained ViT-Small feature extractor paired with a
Transformer-based diffusion denoiser, with all values taken from the
configuration file provided in its official GitHub repository. However, we could not
train DiffLoc on this subsampled dataset, and therefore omit it from
Table~\ref{tab:supp_oxford_results}. The reason is that DiffLoc's
static-object-aware pooling is supervised by semantic static/dynamic masks (originally
produced by an SPVNAS~\cite{tang2020searching} segmentation network), but because the released subsampled point-clouds carry only $(x,y,z)$ coordinates, both the intensity channel and the per-point segmentation labels were unavailable and set to zero. As a result, the segmentation branch received no meaningful supervision, the ground-truth masks were
identically zero, causing its auxiliary loss to remain at zero and leaving DiffLoc
without the static-object cue it normally depends on. For MB-Loc, the configuration reported in Table~\ref{tab:supp_oxford_results} uses $P=8$ Z-slicing planes
at $G=512$, optimized with Adam (learning rate $3\times10^{-3}$, weight decay
$1\times10^{-6}$, step decay $\times0.85$ every $20$ epochs, KL weight $1\times10^{-4}$)
and decoded by the same backend with $\tau=5.0$\,m, $N_{\max}=2{,}000$, and
confidence $p=0.95$.

\subsection{Per-Sequence Quantitative Results}
\label{subsec:supp_oxford_persequence}

We perform this ablation on the configurations $10\times256\times256$,
$15\times256\times256$, and $8\times512\times512$, at RANSAC $\tau = 5.0$\,m,
$N_{\max} = 2{,}000$, and confidence $p = 0.95$; the per-sequence results are reported
in Table~\ref{tab:supp_oxford_persequence}.

\begin{table}[h]
\centering
\caption{Per-sequence localization accuracy of MB-Loc on the subsampled
Oxford~\cite{barnes2020oxford} test sequences across structural configurations
$(P, G, G)$. Default configuration in \textit{italics}.}
\label{tab:supp_oxford_persequence}
\small
\begin{tabular}{|l|c|c|c|c|}
\hline
\textbf{Config $(P,G,G)$} & \textbf{17-14-03-00} & \textbf{15-13-06-37} & \textbf{18-14-14-42} & \textbf{Avg [m/$^{\circ}$]} \\
\hline\hline
$10\times256\times256$ & 3.50/1.85 & 2.78/1.86 & 2.41/1.69 & 2.89/1.80 \\
$15\times256\times256$ &  3.57/1.63         &       2.81/1.62    &  2.34/1.54       &     2.90/1.59      \\
\hline
$\mathit{8\times512\times512}$ & \textit{3.25/1.79} & \textit{2.44/1.71} & \textit{2.23/1.61} & \textit{2.64/1.70} \\
\hline
\end{tabular}
\end{table}

\clearpage

\clearpage

\bibliography{references}

\end{document}